\documentclass[lettersize,journal]{IEEEtran}
\usepackage{cite}
\usepackage{amsmath,amssymb,amsfonts}
\usepackage{algorithmic}
\usepackage{graphicx}
\usepackage{textcomp}
\usepackage{color}
\usepackage{bm}
\usepackage[caption=false,font=normalsize,labelfont=sf,textfont=sf]{subfig}
\def\BibTeX{{\rm B\kern-.05em{\sc i\kern-.025em b}\kern-.08em
    T\kern-.1667em\lower.7ex\hbox{E}\kern-.125emX}}

\usepackage{enumerate}
\usepackage{booktabs}
\usepackage{balance}
\usepackage{cancel}

\usepackage{adjustbox}
\usepackage{gensymb}
\usepackage{color, colortbl}
\definecolor{LightCyan}{rgb}{0.88,1,1}

\allowdisplaybreaks
\newcommand{\Renyi}{R\'{e}nyi }
\newcommand\subscr[2]{#1_{\textup{#2}}}

\newcommand{\accgap}{\subscr{\mathtt{acc}}{gap}}
\newcommand{\accmin}{\mathtt{\subscr{acc}{min}}}
\newcommand{\dpgap}{\subscr{\mathtt{dp}}{gap}}
\newcommand{\eqoddsgap}{\subscr{\mathtt{eqodds}}{gap}}

\newcommand{\ind}{\perp\!\!\!\!\perp} 
\newcommand{\acc}{\mathtt{acc}}
\newcommand{\alphat}{\Tilde{\alpha}}
\makeatletter
\newcommand{\newlineauthors}{%
  \end{@IEEEauthorhalign}\hfill\mbox{}\par
  \mbox{}\hfill\begin{@IEEEauthorhalign}
}
\makeatother

\begin{document}

\title{Classification Utility, Fairness, and Compactness \\ via Tunable Information Bottleneck \\ and  R\'{e}nyi Measures}

\author{Adam~Gronowski,
        William~Paul,
        Fady~Alajaji,
        Bahman~Gharesifard,
        and~Philippe~Burlina
\thanks{This work was supported in part by the Natural Sciences and Engineering Research Council of Canada. \textit{(Corresponding author: Adam Gronowski.)} }%
\thanks{A preliminary version of part of this work was presented 
as~\cite{gronowski2022rfib} in the 2022 Canadian Workshop on Information Theory.}
\thanks{A. Gronowski and F. Alajaji are with the Department of Mathematics
and Statistics, Queen’s University, Kingston, ON K7L 3N6, Canada (e-mail:
adam.gronowski@queensu.ca; fa@queensu.ca).}
\thanks{W. Paul and P. Burlina are with The Johns Hopkins University Applied Physics Laboratory, 11100 Johns Hopkins Road, Laurel, MD 20723, USA (e-mail:
william.paul@jhuapl.edu; philippe.burlina@jhuapl.edu).}
\thanks{B. Gharesifard is with the Electrical \& Computer Engineering Department, University of California at Los Angeles, Los Angeles, CA 90095 USA (e-mail:
gharesifard@ucla.edu).}%
}

\maketitle

\begin{abstract}
Designing machine learning algorithms that are accurate yet fair, not discriminating based on any sensitive attribute, is of paramount importance for society to accept AI for critical applications. In this article, we propose a novel fair representation learning method termed the \Renyi Fair Information Bottleneck Method (RFIB) which incorporates constraints for utility,
fairness, and compactness (compression) of representation, and apply it to image and tabular data classification. A key attribute of our approach is that
we consider – in contrast to most prior work – both demographic
parity and equalized odds as fairness constraints, allowing for
a more nuanced satisfaction of both criteria. Leveraging a variational approach, we show that our objectives yield a loss function involving classical Information
Bottleneck (IB) measures and establish an upper bound in terms of two \Renyi measures of order~$\bm{\alpha}$ on the mutual information IB term measuring compactness between the input and its encoded embedding. We study the influence of the $\bm{\alpha}$ parameter as well as two other tunable IB parameters on achieving utility/fairness trade-off goals, and show that the $\bm{\alpha}$ parameter gives an additional degree of freedom that can be used to control the compactness of the representation. Experimenting on three different image datasets (EyePACS, CelebA, and FairFace) and two tabular datasets (Adult and COMPAS), using both binary and categorical sensitive attributes, we show that on various utility, fairness, and compound utility/fairness metrics RFIB outperforms current state-of-the-art approaches.

\end{abstract}

\begin{IEEEkeywords}
Deep learning, fair representation learning, equalized odds, demographic parity, classification, information bottleneck (IB), \Renyi divergence, \Renyi cross-entropy.
\end{IEEEkeywords}


\section{Introduction}
\IEEEPARstart{M}{achine} learning algorithms are used for a variety of high stake applications such as loan approvals, police allocation, admission of students, and disease diagnosis. In spite of its vast benefits, the use of automated algorithms
that are not designed to also address potential bias and fairly serve members of diverse groups can lead to harm and exacerbate social inequities. The problem of developing algorithms that are both accurate and \emph{fair}, i.e., algorithms that do not discriminate against individuals because of their gender, race, age, or other sensitive (protected) attributes, has now become paramount to the effective deployment of production-grade AI systems that could be accepted and adopted by society as a whole.

Fair machine learning methods have been developed for multiple domains such as automated healthcare diagnostics and treatments delivery~\cite{kinyanjui2020fairness}, natural language processing~\cite{bolukbasi2016man}, finance~\cite{friolo2022cryptographic}, and others~\cite{zemel2013learning,prost2019toward,kinyanjui2020fairness,caton2020fairness,albiero2021gendered}. One way to create such fair machine learning methods is through learning \emph{fair representations} that can be used with existing architectures.
These representations would allow for making accurate predictions while ensuring fairness. However, this leads to the difficulty that developing fair representations may involve trade-offs and is further complicated by the existence of various metrics for measuring fairness outcomes, often tailored towards different applications and settings. 

Our work entails the development of a fair representation learning method that addresses a number of trade-offs. We consider the case where the sensitive attributes are directly accessible, and for the majority of experiments focus on the case where bias is caused by a lack of data for a protected subgroup.
Unlike most prior studies that tend to focus on satisfying a single type of fairness constraint, we consider here how to jointly address and balance two of the arguably most important and widely used definitions for fairness, \emph{demographic parity} and \emph{equalized odds} \cite{zemel2013learning, hardt2016, zhao2019conditional}.
We also examine different classical trade-offs between fairness and utility (commonly measured via accuracy). Trade-offs between utility and fairness arise as a result of interventions on models or data to make models more fair, which may yield decreased bias, but may also result in affecting utility. We study how these trade-offs are impacted by  ``compactness.'' More specifically, we develop a variational approach taking into account different information-theoretic metrics that balance the above two constraints on fairness with utility and compactness. We also show how to analytically simplify the resulting loss function and relate it
to the {\em Information Bottleneck (IB)} principle \cite{tishby2000information}, and then exploit bounds to compute these metrics.

This work thus makes the following 
novel contributions:
\begin{enumerate}

\item We develop a novel variational method that balances a triplet of objectives, consisting of utility (accuracy),  fairness (itself balancing two types of fairness constraints), and compression/compactness. Specifically, in contrast to prior work on fairness which narrowly uses either the demographic parity or equalized odds constraints, our loss includes both types of constraints. We relate analytically the resulting loss function to the classical IB method.  
 
\item Operationally, we derive a simple upper bound on the mutual information between the data and its representation in terms of the \Renyi divergence~\cite{renyi} and the \Renyi cross-entropy \cite{Alba, bhatia2021least,thierrin2022} of order $\alpha$. We study the effect of this added flexibility on achieving a balance between fairness and accuracy. 
 
\item Using various datasets such as CelebA, EyePACS, FairFace, Adult, and COMPAS, and working with both binary and categorical sensitive variables, we compare with methods of record that intervene on the model or methods that intervene on training data. We show that  our method overall performs best. We establish these comparisons via a number of metrics that measure utility and fairness individually or in a combined metric, including two different types of fairness constraints.
\end{enumerate}

The rest of the article is organized as follows. We present related work in Section~II and derive a cost function for our method in Section~III. In Section~IV, we describe the details of our implementation, describe the metrics and datasets used, and present extensive experimental results. Finally, we conclude the article with Section~V.

\section{Related Work}


\subsection{Fairness Approaches}

We summarize existing work on fair machine learning in three categories. For a broad-strokes categorization of fairness approaches, one can think along the lines of {\em interventions} made either on: 1) the model output, 2) the training data, or 3) the model itself. Each of these are motivated by different inductive biases.

\subsubsection{Interventions on Model Outputs Including Recalibration and Thresholding}
The inductive bias here is that irrespective of the cause of bias, fairness can be addressed at the output of the model. Some methods adjust the decision threshold to achieve equal odds across groups, as in~\cite{hardt2016}, or re-calibrate the model for different subpopulations, as in~\cite{pleiss2017fairness}. These methods work well for categorical data, but they do not address the underlying causes of bias in the data and the model, which may be more important for image/video applications, as argued in~\cite{hardt2016}. Therefore, we explore a different approach.

\subsubsection{Data Interventions: Generative Models and Style Transfer} The idea here is to modify the data to reduce bias since data imbalance can lead to biased models. Methods that alter the training data can censor sensitive information, blind the model to protected factors, augment the data, or re-balance the data by re-sampling or re-weighting. Data augmentation methods use generative models or style transfer/image translation. Some examples of generative models are based on generative adversarial networks (GANs)~\cite{karras2019style,grover2019bias,paul2020tara}, variational autoencoders~\cite{hwang2020fairfacegan,sattigeri2018fairness}, or latent space optimization~\cite{paul2020tara}. An example of image translation is~\cite{quadrianto2019discovering}. However, these methods may have limitations in preserving the diversity of styles within subpopulations, which may be important for image/video applications. These methods are also related to domain adaptation and distributional shift scenarios.

Generating data with these methods has drawbacks as it may be hard to isolate and manipulate the protected factors without affecting other features (a problem called {\em entanglement}). Generative methods may also introduce artifacts in synthetic images, which could lower the performance and utility of the debiased models without significantly improving fairness. Based on these observations,~\cite{paul2020tara} suggests that model interventions may be better than data interventions using generative models. These limitations motivate our approach, which belongs to the category of model interventions, discussed next.
 
\subsubsection{Interventions on Models via Adversarial and Variational Approaches}
Such methods are grounded on the inductive bias that dependence of the prediction on protected factors is a cause of lack of fair predictions; as a result, these methods generally aim to remedy this dependence at the encoding of the data, rendering them blind to protected factors.

 Studies such as~\cite{wadsworth2018achieving} and
 \cite{zhang2018mitigating} use an adversarial network to penalize the prediction network if it could predict a protected factor. 
Other studies employ adversarial representation learning to remove protected factor information from latent representations, including \cite{edwards2015censoring,beutel2017data,madras2018learning,roy2019mitigating,zhao2019conditional}.
%
Applications using this principle include~\cite{pfohl2019creating} which uses adversarial learning to develop fair models of cardiovascular disease risk, while \cite{gitiaux2021learning} explores the statistical properties of fair representation learning and \cite{grari2019fairness} applies an adversarial approach for continuous features. 
Similar to our work, \cite{song2019learning} and the more general recent work in~\cite{kairouz2022generating} employ an information-theoretic approach to learn fair representations. More specifically in~\cite{kairouz2022generating}, 
the authors investigate an adversarial censoring method to generate universal fair representations that are least informative about sensitive attributes given a utility constraint.
In contrast to the above techniques, our method does not use adversarial training.

In addition to the methods mentioned above, there exist non-adversarial methods that modify the model via incorporation of multiple constraints in a variational setting. Many of these are closely related to the IB method that we discuss next.

\subsection{Information Bottleneck Methods}
The IB method, originally proposed by Tishby \textit{et al.} \cite{tishby2000information}, is a method that seeks to develop representations that are both compact and expressive by minimizing and maximizing two mutual information terms. Alemi \textit{et al.} first developed a variational approximation of the IB method by parameterizing it using neural networks and this was followed by multiple variations such as the nonlinear information bottleneck \cite{kolchinsky2019nonlinear} and conditional entropy bottleneck \cite{fischer2020conditional}. Many recent generalizations of the IB method have been developed including  \cite{makhdoumi2014information, strouse2017deterministic,hsu2018generalizing,asoodeh2020bottleneck,weng2021information} while \cite{goldfeld2020information,zaidi2020information} investigated its connections to deep learning theory and privacy applications.

Techniques related to the information bottleneck have been used for fair representation learning. First proposed by \cite{zemel2013learning}, fair representation learning consists of mapping input data to an intermediate representation that remains informative but discards unwanted information that could reveal the protected sensitive factors. This is related to the IB problem and several works have explored the connection between the two, such as 
~\cite{ghassami2018fairness} and~\cite{skoglund2020}, 
where fair representations are acquired through the minimization and maximization of various mutual information terms.

Our work is closest to~\cite{skoglund2020} but we depart from it in several important ways, including through the use of a more general loss formulation, satisfying broader constraints of fairness, classifying images rather than focusing solely on tabular data, and entailing the use of R\'{e}nyi divergence and \Renyi cross-entropy \cite{Alba,bhatia2021least,thierrin2022}. To our knowledge we are the first to use R\'{e}nyi divergence and cross-entropy for fair representation learning but there have been several recent works based on \Renyi information measures and their variants. These include an IB problem under a \Renyi entropy complexity constraint \cite{weng2021information}, bounding the generalization error in learning algorithms~\cite{esposito20}, \Renyi divergence variational inference~\cite{varInf}, \Renyi differential privacy \cite{mironov2017renyi} and the analysis and development of GANs \cite{bhatia2021least, rgan, cumulant_gan, kurri2021realizing, kurri2022, welfert2023alphadalphaggans}. In addition, Baharlouei \textit{et al.} \cite{baharlouei2019r} developed a fair representation method but one based on \Renyi correlation rather than divergence.

\subsection{Connections to Other Work}
Fairness is related to the problem of domain adaptation (DA)~\cite{ben2010theory,ganin2016domain,long2015learning, zhang2021multiple} that consists of training a neural network on a source dataset to obtain good accuracy on a target dataset that is different from the source. This is especially true for the case of a severe data imbalance that we consider in this work where training data is completely missing for a protected subgroup. In this case, achieving fairness is similar to the DA problem of improving performance on a complete target dataset that includes all groups after training on an incomplete source dataset.

Our work is also related to group distributionally robust optimization (GDRO) methods \cite{sagawa2019distributionally,sohoni2020no} that address the problem of performance disparity among different subgroups by minimizing the worst-case loss among different subgroups. Reducing these accuracy differences among subgroups is something our method also addresses, but while this is the sole objective for GDRO methods, we consider this problem in relation to multiple other fairness criteria.

There are other studies about finding a balance between accuracy and fairness such as Zhang \textit{et al.} \cite{zhang2021balancing}. However, they achieved this objective by finding early stopping criteria rather than through a fair representation preprocessing method like we use here; also, unlike our method, finding a balance between multiple fairness constraints is not investigated. 

Finally, there are hybrid methods, such as the one proposed by Paul \textit{et al.} in~\cite{paul2020tara}, that use a combination of the previously discussed techniques of interventions on data, model, or output.  
However, such techniques are quite limited in scope compared to the more varied objectives considered here which involve jointly achieving utility, fairness, and compactness.

\section{Methods}
We take an information-theoretic approach to fairness (refer to \cite{coverThomas,alajaji} for background knowledge on standard information measures such as Shannon entropy and mutual information). We represent input data as a random variable $X \in \mathcal{X}$ and sensitive information as a random variable $S \in \mathcal{S}$. Our goal is to use the data $X$ to predict a target $Y \in \mathcal{Y}$ but in a way that is uninfluenced by the sensitive information $S$. To reach this objective, we adopt a variational approach, which we call {\em R\'enyi Fair Information Bottleneck} (RFIB), to encode the data into a new representation $Z \in \mathcal{Z}$. The representation can then be used with existing deep learning model architectures to draw inferences about $Y$. As we compute $Z$ from $X$ without accessing data from $Y$ or $S$, we can make a standard assumption similar to \cite{deepvariational} and assume that the Markov chain 
$$(Y,S) \rightarrow X \rightarrow Z \rightarrow \hat{Y}$$
holds, where $\hat{Y}$ is the prediction of $Y$ based on $Z$ ($\hat{Y}$ is a function of $Z$). For simplicity, we assume in this section that all random variables are discrete, though a similar derivation holds for a mix of continuous and discrete random variables.

\subsection{Fairness Defined}
Among the three principal definitions of fairness -- \emph{demographic parity}, \emph{equalized odds}, and \emph{equality of opportunity} -- we focus on addressing both demographic parity and equalized odds since a) equalized odds is a stronger constraint than equality of opportunity (equalized odds requires both an equal true positive rate and equal false positive rate across the sensitive variable, whereas equality of opportunity only requires an equal false positive rate), and b) demographic parity, also called statistical parity, is an altogether different type of constraint compared to the former two constraints in that the requirement of independence does not involve the actual target label value.

For demographic parity, the goal is for the model's prediction $\hat{Y}$ to be independent of the sensitive variable $S$\cite{zemel2013learning}, i.e.,
\begin{equation}\label{eq:dp}
P(\hat{Y}=\hat{y})=P(\hat{Y}=\hat{y} | S=s),
\end{equation}
for all $s,\hat{y}$,
while for equalized odds the goal is to achieve this independence by conditioning on the actual target $Y$\cite{hardt2016}, i.e.,
\begin{equation}\label{eq:eqodds}
P(\hat{Y}=\hat{y}| Y=y)=P(\hat{Y}=\hat{y}| S=s,Y=y)
\end{equation}
for all $s,\hat{y},y$.

\subsection{Lagrangian Formulation}
We formulate a Lagrangian by minimizing and maximizing various mutual information terms, similar to \cite{du2012privacy,deepvariational}. To encourage equalized odds, we want as per \eqref{eq:eqodds} that $\hat{Y}$ and $S$ are conditionally independent given $Y$ (written as $(\hat{Y} \ind S) \mid Y$); this is equivalent to requiring that  $I(\hat{Y} ; S | Y)=0$.
To achieve this objective, we minimize the proxy measure $I(Z ; S |Y)$ over $P_{Z \mid X}$. This is justified by the following:
by the chain rule for mutual information, we can write
\begin{align}
I(Z, \hat{Y} ; S | Y)  &=I(Z ; S | Y)+ \underbrace{I(\hat{Y} ; S | Z, Y)}_{=0} \nonumber \\
 &=I(\hat{Y} ; S | Y)+\underbrace{I(Z ; S |\hat{Y}, Y)}_{\geqslant 0}, \label{eq:justification}
\end{align}
 where $I(\hat{Y} ; S | Z, Y)=0$, since by the Markov chains $(Y, S) \rightarrow Z \rightarrow \hat{Y}$ and $Y \rightarrow Z \rightarrow \hat{Y}$,\footnote{Note that these two Markov chains are implied by the original Markov chain formulation $(Y,S) \to X \to Z \to \hat{Y}$.} we have for all $\hat{y},z,s,y$ that 
 \begin{align*}
     P(\hat{Y}=\hat{y}| Z=z, S=s, Y=y) 
     &= P(\hat{Y}=\hat{y} | Z=z) \\
     &= P(\hat{Y}=\hat{y}|Z=z,Y=y)
 \end{align*}
 and hence $(\hat{Y} \ind S) \mid (Z, Y) \Leftrightarrow I(\hat{Y} ; S | Z, Y)=0.$ We therefore obtain from (\ref{eq:justification}) that $$0 \leq I(\hat{Y} ; S | Y) \leq I(Z ; S | Y).$$ So by minimizing $I(Z;S| Y)$ over $P_{Z\mid X},$ we are squeezing $I(\hat{Y};S| Y)$ closer to zero. In the ideal case where we can drive $I(Z;S| Y)$ to zero, we readily obtain that $$I(\hat{Y} ; S | Y)=0 \Leftrightarrow (\hat{Y} \ind S) \mid Y$$ as ultimately desired in equalized odds.

To both obtain good classification accuracy (utility) and help promote demographic parity, we maximize $I(Z;Y|S)$. For demographic parity, we want as per \eqref{eq:dp} that $\hat{Y} \ind S$ $\Leftrightarrow I(\hat{Y} ; S)=0,$ while for utility we want high $I(Y;\hat{Y})$ which we address by maximizing $I(Z;Y).$  To achieve $I(\hat{Y} ; S)=0$ we require $I(Z ; S)=0 \text { (i.e., } Z \ind S)$ since 
$$0 \leq I(S ; \hat{Y}) \leq I(S ; Z)$$ 
by the Data Processing Inequality~\cite{coverThomas, alajaji} as $S \rightarrow Z \rightarrow \hat{Y}$ holds. Now by the mutual information chain rule, we can write
\begin{align} I(Z ; Y, S) & =I(Z ; Y)+I(Z ; S | Y) \nonumber \\ & =I(Z ; S)+I(Z ; Y | S).\end{align} 
In the ideal condition where $I(Z;S \mid Y) = 0$ (i.e., equalized odds is achieved), a necessary condition for $I(Z;S)=0$ is that $I(Z;Y| S) = I(Z;Y)$ which we want to maximize. Thus maximizing $I(Z;Y| S)$ promotes both demographic parity and high utility.



Finally, we minimize $I(Z;X|S,Y)$, a compression term similar to one from the IB problem~\cite{deepvariational}. This minimization further encourages $Z$ to discard information irrelevant for drawing predictions about $Y,$ hence improving generalization capability and reducing the risk of keeping nuisances (irrelevant data that can degrade performance).

Combining all above mutual information terms leads to a Lagrangian, ${\cal L}$, that we seek to minimize over the encoding conditional distribution $P_{Z|X}$. The Lagrangian is given by
\begin{align}
\cal{L} &= I(Z;S|Y) + I(Z;X|S,Y) -\lambda_1 I(Z;Y) 
    -\lambda_2 I(Z;Y|S),
    \label{eq:lagran1}
\end{align}
where $\lambda_1$ and $\lambda_2$ are hyper-parameters.
Developing this Lagrangian, we have that
\begin{align}
\cal{L} &= H(Z|Y) - H(Z|S,Y) +H(Z|S,Y) \nonumber \\&\quad- H(Z|X,S,Y) - \lambda_1I(Z;Y) - \lambda_2 I(Z;Y|S)  \nonumber \\
& = H(Z|Y) - H(Z|X) -  \lambda_1I(Z;Y) - \lambda_2 I(Z;Y|S) \nonumber \\
\begin{split}
&= H(X) - H(Z,X) - [H(Y) - H(Z,Y)]\\&\quad -\lambda_1I(Z;Y) - \lambda_2 I(Z;Y|S) \nonumber \\
\end{split}
\\
&= I(Z;X) - I(Z;Y) - \lambda_1I(Z;Y) - \lambda_2 I(Z;Y|S) \nonumber \\
&= I(Z;X) - (\lambda_1 + 1)I(Z;Y) - \lambda_2 I(Z;Y|S), 
\end{align}
where $H(\cdot)$ denotes entropy, and the second equality follows from the Markov chain assumption $(Y,S) \rightarrow X \rightarrow Z$. Hence, we have shown that the Langrangian $\mathcal{L}$ admits a simpler equivalent expression given by
\begin{equation}\label{simple-L}
\mathcal{L} = I(Z;X) -\beta_1 I(Z;Y) -\beta_2 I(Z;Y|S),
\end{equation}
where $\beta_1 = \lambda_1 + 1 \ge 1$ and $\beta_2 = \lambda_2 \ge 0$.
This~simpler Lagrangian is easier to compute while exactly maintaining the properties of the original one. It also reveals a direct relation of the original Lagrangian with the first two terms being exactly equivalent to the ``classical IB'' formulation. Note that both the $I(Z;X)$ term and $I(Z;Y|S)$ term contribute to fairness. Minimizing $I(Z;X)$ results in increased compression which can lead to increased fairness at the possible expense of accuracy. The $I(Z;Y|S)$ term contributes to both fairness through the conditioning on $S$ and to accuracy through the maximization of the information between $Z$ and $Y.$ The hyper-parameters $\beta_1$ and $\beta_2$ control trade-offs between accuracy (or ``utility'') and fairness, as a higher weight on $\beta_1$ and $\beta_2$ reduces the influence of the $I(Z;X)$ compression term.
As $I(Z;Y)$ is partially derived from the $I(Z;S|Y)$ term designed to improve equalized odds, using a higher $\beta_1$ over $\beta_2$ should give more priority to improving equalized odds, whereas a higher $\beta_2$ should result in improved demographic parity. This allows for more nuanced outcomes compared to other methods that focus rigidly on a single fairness metric. It is also possibly an interesting tool for policy makers to translate those more balanced and nuanced versions of fairness into an ``engineered system.''

\subsection{Variational Bounds} We use a variational approach to develop bounds on the three terms in the Lagrangian $\mathcal{L}$ in \eqref{simple-L}, finding lower bounds for the terms to be maximized and an upper bound for the term to be minimized. The Markov chain property $(Y,S) \rightarrow X \rightarrow Z$ results in the joint distribution $P_{SYXZ}$ factoring as as $P_{SYX}P_{Z|X}.$ 

The distribution $P_{Z|X}$ is a parametric stochastic encoder to be designed while all other distributions are fully determined by the joint data distribution $P_{S,X,Y}$, the encoder, and the Markov chain constraint. To simplify notation, we simply write $P_{Z|X}$ rather than including the parameter $P_{Z|X,\theta}$, with $\theta$ denoting network weights. Computing the mutual information terms requires the usually intractable or difficult to compute distributions $P_{Y|S,Z}$, $P_{Y|Z}$, and $P_{Z}$ \cite{kingma2013auto,deepvariational}; we thus replace them with variational approximations $Q_{Y|S,Z}$, $Q_{Y|Z}$ and $Q_{Z}$, respectively. We next derive a simple upper bound for the compression term $I(Z;X)$ with the novel use of R\'{e}nyi's divergence and cross-entropy of order~$\alpha$, which provide a tunable extra degree of freedom for the variational version of $\mathcal{L}$: 
\begin{align}
I(Z ; X) &= \min_{Q_Z} D_{KL}\left(P_{XZ} \| P_X Q_Z\right) \nonumber \\
& \leq E_{P_X} D_{KL}\left(P_{Z \mid X} \| Q_Z\right) \nonumber \\
& \leq \begin{cases}E_{P_X} D_\alpha\left(P_{Z \mid X} \| Q_Z\right) & \text { if } \alpha\geq1  \\
E_{P_X} H_\alpha\left(P_{Z \mid X} ; Q_Z\right) & \text { if } \alpha<1.\end{cases} \label{eq:bound1}
\end{align}
The first identity follows from the non-negativity of the Kullback-Leibler (KL) divergence (see for example~\cite[Problem~2.18]{alajaji},\cite{deepvariational,song2019learning,skoglund2020}).

For the case of $\alpha \ge 1$ in the final inequality given in~\eqref{eq:bound1}, we take the \Renyi divergence $D_\alpha(\cdot \| \cdot)$ of order $\alpha$ (e.g., see~\cite{renyi,renyiDivergence}), rather than the KL divergence as typically done in the literature, where
\begin{equation}
    D_\alpha(P\|Q) = \frac{1}{\alpha-1} \log \left(\sum_{x\in \mathcal{X}} P(x)^{\alpha}Q(x)^{1-\alpha}  \right)
\end{equation}
for $\alpha>0$, $\alpha \ne 1$ and distributions $P$ and $Q$ with common support~$\mathcal{X}$.\footnote{When $P$ and $Q$ are both probability density functions, then $D_\alpha(P||Q) = \frac{1}{\alpha-1} \log \left( \int_{\mathcal{X}} P(x)^{\alpha}Q(x)^{1-\alpha} \, dx \right)$.}
By the continuity property of $D_\alpha$ in $\alpha$, see~\cite{renyiDivergence}, we define its extended orders at $\alpha=1$ and $\alpha=0$ as
\begin{align}
&D_1(P\|Q) := \lim_{\alpha \to 1} D_\alpha(P\|Q)= D_{KL}(P\|Q)
\end{align}
and
\begin{align}
&\hspace{-0.1in} D_0(P\|Q) :=\lim_{\alpha \to 0} D_\alpha(P\|Q) = -\log Q(x: P(x)>0), \label{extended-orders}
\end{align}
which is equal to 0 when $P$ and $Q$ share a common support.
This upper bound in~\eqref{eq:bound1} readily holds for $\alpha \ge 1$ since $D_\alpha$ is non-decreasing in $\alpha$ and
$D_1(P\|Q) = D_{KL}(P\|Q)$.

For the case of $\alpha < 1$ in~\eqref{eq:bound1}, we take the \Renyi cross-entropy $H_\alpha(\cdot ; \cdot)$ of order $\alpha$, used recently in \cite{bhatia2021least } to generalize the original loss function of GANs \cite{goodfellow2014generative}:\footnote{When $P_{Z|X=x}$ and $Q_Z$ are both probability density functions, then the (differential) \Renyi cross-entropy is given by $h_\alpha\big(P_{Z | X=x} ; Q_Z\big)= 
\frac{1}{1-\alpha} \log \big(\int_{\mathcal{Z}} P_{Z | X}(z | x) Q_Z(z)^{\alpha-1} \, dz\big).$}
\begin{align}
H_\alpha\big(P_{Z | X=x} ; Q_Z\big)= 
\frac{1}{1-\alpha} \log \Big(\sum_{z\in \mathcal{Z}} P_{Z | X}(z | x) Q_Z(z)^{\alpha-1}\Big).
\end{align}
We hence justify the inequality in \eqref{eq:bound1} when $\alpha < 1$ as follows:
\begin{align}
E_{P_X} D_{KL}\left(P_{Z|X} \| Q_Z \right)& 
=E_{P_X} \sum_{z\in \mathcal{Z}} P_{Z|X}(z|x) \log \frac{P_{Z|X}(z|x) }{Q_Z(z)} \nonumber \\
=&E_{P_X}\underbrace{\left[\sum_{z\in \mathcal{Z}} P_{Z|X}(z|x) \log \frac{1}{Q_Z(z)}\right]}_{=H(P_{Z|X=x};Q_Z)} \nonumber \\
&\hspace{-0.15in} + E_{P_X}\underbrace{\left[\sum_{z\in \mathcal{Z}} P_{Z|X}(z|x) \log P_{Z|X)}(z|x)\right]}_{=-H(Z|X=x)\leq 0} \nonumber \\
\leq& E_{P_X}H(P_{Z|X=x};Q_Z) \nonumber \\
\leq& E_{P_X}H_\alpha(P_{Z|X=x};Q_Z), \label{cross-derivation}
\end{align}
where $H(\cdot;\cdot)$ is the Shannon cross-entropy and the final inequality follows since R\'{e}nyi's cross-entropy is non-increasing in $\alpha$ and  $\lim _{\alpha \rightarrow 1} H_\alpha\left(P_{Z \mid X=x};Q_Z\right)=H\left(P_{Z|X=x} ; Q_Z\right)$ \cite{bhatia2021least}.

\smallskip
{\em Remark~1}:
\label{rem1} The above derivation is carried for discrete distributions. It still holds for continuous distributions except that for the first inequality in~\eqref{cross-derivation}, we need to ensure that the conditional differential entropy $h(Z|X=x)$ is non-negative. More specifically, if $P_{Z|X=x}$ is an uncorrelated Gaussian vector (which we use below in the derivation of the R\'{e}nyi divergence and ensuing experiments), we ensure the non-negativity of the vector's conditional differential entropy by selecting large enough variance components $\sigma_i^2$ (see Footnote~\ref{footnote4}). This also guarantees the non-negativity of the differential Shannon and \Renyi cross-entropies: $h(P_{Z|X=x};Q_Z)$ and $h_\alpha(P_{Z|X=x};Q_Z)$ with $\alpha<1$.

{\em Remark 2}: If we take $\alpha <1$ for the \Renyi divergence or $\alpha \ge 1$ for the \Renyi cross-entropy in~\eqref{eq:bound1}, we 
no longer have an upper bound on $I(Z;X)$. But in these cases,  the terms in~\eqref{eq:bound1} can be considered as approximations to the compression term $I(Z;X)$ that are $\alpha$-tunable. In other words, as the \Renyi divergence is non-decreasing in $\alpha$ and the \Renyi cross-entropy is non-increasing in $\alpha$, choosing values of $\alpha < 1$ or $\alpha \geq 1$, respectively, will result in a smaller approximation for $I(Z;X)$. This allows us to fully control the weight we put on that term compared to the others in the cost function in~\eqref{simple-L} by adjusting $\alpha$, being able to adjust it all the way down to zero if desired.

We can similarly leverage the non-negativity of KL divergence to get lower bounds on $I(Z;Y)$ and $I(Z;Y|S)$:
\begin{equation}
I(Z;Y) \geq \mathbb{E}_{P_{Y,Z}} \left[ \log Q_{Y | Z}(Y | Z)\right]+H(Y),
\end{equation}
\begin{equation}
\hspace{-0.09in} I(Z;Y|S) \geq \mathbb{E}_{P_{S,Y,Z}} \hspace{-0.03in} \big[\log Q_{Y | S,Z }(Y |S,Z)\big]+H(Y|S).
\end{equation}
As the entropy $H(Y)$ and conditional entropy $H(Y|S)$ of the labels do not depend on the parameterization they can be ignored for the optimization. 

\subsection{Computing the Bounds}\label{compute-bounds}
To compute the bounds in practice we use the reparameterization trick \cite{kingma2013auto}. Modeling $P_{Z|X}$ as a density, we let $P_{Z|X}dZ$ = $P_E dE$, where $E$ is a Gaussian (normal) random variable with zero mean and unit variance and $Z=f(X,E)$ is a deterministic function, allowing us to backpropagate gradients and optimize the parameter via gradient descent. We use the data's empirical densities to estimate  
$P_{X,S}$ and $P_{X,Y,S}$.

Considering a batch $D = \{x_i,s_i,y_i \}_{i=1}^N$ this finally leads to the following RFIB cost function to minimize:
\begin{equation}\label{final-loss}
\begin{split}
    J_\text{RFIB} &= \frac{1}{N}\sum_{i=1}^{N} \Big[ F_{\alpha}(P_{Z|X=x_i},Q_Z) \\&\quad- \beta_1 \mathbb{E}_E\left[\log \left(Q_{Y|Z}\left(y_i|f(x_i,E)\right)  \right)  \right]\\ &\quad- \beta_2 \mathbb{E}_E\left[\log \left(Q_{Y|S,Z}\left(y_i|s_i,f(x_i,E)\right)  \right)  \right]\Big],
\end{split}
\end{equation}
where we estimate the expectation over $E$ using a single Monte Carlo  sample (as in~\cite{deepvariational}) and

$$F_\alpha(P_{Z|X=x_i},Q_Z)= \begin{cases}
D_\alpha(P_{Z|X=x_i}\|Q_Z) & \text{if $\alpha\ge 1$} \\
H_\alpha(P_{Z|X=x_i};Q_Z) & \text{if $\alpha < 1$}. 
\end{cases} $$

We will compare the performance of our RFIB system with both the \emph{IB} \cite{deepvariational} and the \emph{conditional fairness bottleneck} (CFB) \cite{skoglund2020} schemes.
Note that if we set $\alpha = 1$ and $\beta_2 = 0$ in~\eqref{simple-L}, or in the variational loss function~\eqref{final-loss}, we recover the IB system.
Furthermore, setting $\alpha = 1$ and forcing $\beta_1=0$ (by relaxing its range) yields the CFB loss function. We will therefore use the parameters pairs of $(\alpha=1,\beta_1=0)$ and $(\alpha=1,\beta_2=0)$ as corresponding to the CFB and IB systems, respectively.
\subsection{Derivation of $D_{\alpha}\left(P_{Z \mid X} \| Q_{Z}\right)$ for Gaussians}\label{derivation-div}


Here we calculate the R\'enyi divergence term of our cost function when $P_{Z|X}$ and $Q_Z$ are multivariate Gaussian distributions with $P_{Z|X}$ having a diagonal covariance structure while $Q_Z$ being a spherical Gaussian. More specifically, we~let 
\begin{align*}
P_{Z|X}&= \mathcal{N}\left(Z|\mu_{\text enc}(X),\text{diag}(\sigma_{\text enc}^2(X)) \right) \\ 
Q_Z &= \mathcal{N}(Z|\underbar{0},\gamma^2I_d),
\end{align*}
where $\mu_{\text enc}$ and $\sigma_{\text enc}^2$ are $d$-dimensional mean and variance vectors (that depend on $X$), $\text{diag}(\sigma_{\text enc}^2(X))$ is a $d\times d$ diagonal matrix with the entries of $\sigma_{\text enc}^2$ on the diagonal, $\underbar{0}$ is the all-zero vector of size $d$, $\gamma$ is a positive scalar, and $I_d$ is the $d$-dimensional identity matrix. For simplicity of notation, in the rest of the article we write $\mu_{\text enc}(X)$ as $\mu_{\text enc}$ and $\text{diag}(\sigma_{\text enc}^2(X))$ as $\sigma^2_{\text enc}I_d$.

Starting with the closed-form expression of the R\'enyi divergence of order $\alpha$ ($\alpha>0$, $\alpha\neq 1$) derived in \cite{gil2013renyi,burbea1984convexity}, we have
\begin{align}
D_{\alpha}\left(P_{Z|X}\| Q_Z\right)  =& \frac{\alpha}{2}\left(\mu_{\text enc}'{[\left(\Sigma_{\alpha}\right)^{*}}]^{-1}\mu_{\text enc} \right) \nonumber \\ &
-\frac{1}{2(\alpha-1)} \ln \frac{\left|\left(\Sigma_{\alpha}\right)^{*}\right|}{\left|\sigma^2_{\text enc}I_d\right|^{1-\alpha} |\gamma^2 I_d|^{\alpha}} 
\end{align}
where $\mu_{\text enc}'$ is the transpose of $\mu_{\text enc}$, \[
\left(\Sigma_{\alpha}\right)^{*}=\alpha \gamma^2 I_d+(1-\alpha) \sigma^2_{\text enc}I_d,
\]
and 
$$
\alpha[\sigma^2_{\text enc}I_d]^{-1} + (1-\alpha)[\gamma^2 I_d]^{-1}
$$
is positive definite.
Then 
\begin{align}
\frac{\alpha}{2}\left(\mu_{\text enc}'[\left(\Sigma_{\alpha}\right)^{*}]^{-1}\mu_{\text enc} \right) = \frac{\alpha}{2}\sum_{i=1}^d \frac{ \mu_i^2}{\alpha \gamma^2+(1-\alpha)\sigma_i^2},  
\end{align}
where $\mu_i$ and $\sigma_i$ are the $i$th components of $\mu_{\text enc}$ and $\sigma_{\text enc}$ and
\begin{align}
&\quad\frac{1}{2(\alpha-1)} \ln \frac{\left|\left(\Sigma_{\alpha}\right)^{*}\right|}{\left|\sigma^2_{\text enc}I_d\right|^{1-\alpha} |\gamma^2 I
_d|^{\alpha}  }  \nonumber \\
&= \frac{1}{2(\alpha-1)} \ln \frac{\left|\begin{array}{ccc}
\alpha\gamma^2 + (1-\alpha)\sigma_1^2 & \cdots & 0 \nonumber\\
\vdots & \ddots & \vdots \\
0 & \cdots & \alpha\gamma^2+(1-\alpha)\sigma_d^2
\end{array}\right|}{\left|\begin{array}{ccc}
\sigma_1^2 & \cdots & 0 \\
\vdots & \ddots & \vdots \\
0 & \cdots & \sigma_d^2
\end{array}\right|^{1-\alpha} \left|\begin{array}{ccc}
\gamma^2 & \cdots & 0 \\
\vdots & \ddots & \vdots \\
0 & \cdots & \gamma^2
\end{array}\right|^{\alpha}     } \\
&= \frac{1}{2(\alpha-1)} \ln \frac{\prod_{i=1}^d \left[\alpha\gamma^2 + (1-\alpha)\sigma_i^2\right]}{(\prod_{i=1}^d \sigma_i^2)^{1-\alpha}(\prod_{i=1}^d \gamma^2)^{\alpha} } \nonumber\\
&= \frac{1}{2(\alpha-1)} \sum_{i=1}^d \ln \frac{\alpha\gamma^2+(1-\alpha)\sigma_i^2}{\sigma_i^{2(1-\alpha)}\gamma^{2\alpha}},
\end{align}
yielding that 
\begin{align}
D_{\alpha}\left(P_{Z|X}\| Q_Z\right)&=\frac{\alpha}{2}\sum_{i=1}^d \frac{\mu_i^2}{\alpha \gamma^2+(1-\alpha)\sigma_i^2} \nonumber \\
&\quad-\frac{1}{2(\alpha-1)} \sum_{i=1}^d \ln \frac{\alpha\gamma^2+(1-\alpha)\sigma_i^2}{\sigma_i^{2(1-\alpha)}\gamma^{2\alpha}}. \label{eq:renyiDivergence}
\end{align}
Since we require the matrix $\alpha[\sigma^2_{\text enc}I_d]^{-1} + (1-\alpha)[\gamma^2 I_d]^{-1}$ to be positive definite, the above \Renyi divergence expression is valid for
$$\alpha\gamma^2+(1-\alpha)\sigma_i^2 > 0, \qquad i=1,\ldots,d, $$
or equivalently (recalling that $\alpha>0$, $\alpha\neq 1$) for

\begin{align} \nonumber
0 < \alpha < 1,  \ \sigma_i^2 > 0,& \qquad  i=1,\ldots,d, \\
\alpha > 1,  \sigma_i^2 < \frac{\alpha\gamma^2}{\alpha-1},& \qquad i=1,\ldots,d.  \label{eq:condition}
\end{align}


We finish this section with a remark.

\emph{Remark 3} (\emph{Limit as $ \alpha \rightarrow 1$:}) Taking the limit of the \Renyi divergence as $\alpha \rightarrow 1$ (or setting $\alpha=1$ as an extended order) we recover, as expected, the KL divergence expression between Gaussians:
\begin{align}
&\lim _{\alpha \rightarrow 1} D_{\alpha}(P_{Z|X} \| Q_Z) \nonumber \\
&= \lim _{\alpha \rightarrow 1} \Bigg[ \frac{\alpha}{2}\sum_{i=1}^d\frac{ \mu_i^2}{\alpha\gamma^2+(1-\alpha)\sigma_i^2} \nonumber\\&\quad \qquad -\frac{1}{2(\alpha-1)} \sum_{i=1}^d \ln \frac{\alpha\gamma^2+(1-\alpha)\sigma_i^2}{\sigma_i^{2(1-\alpha)}\gamma^{2\alpha}}\Bigg] \nonumber\\
&= \frac{1}{2} \sum_{i=1}^d \frac{\mu_i^2}{\gamma^2} 
- \lim _{\alpha \rightarrow 1} \frac{1}{2}\Bigg[\sum_{i=1}^d\frac{\gamma^2-\sigma_i^2}{\alpha\gamma^2+(1-\alpha)\sigma_i^2} \nonumber  \ \nonumber \\
& \qquad \qquad \qquad \qquad \qquad +2\ln\sigma_i-\ln\gamma^2\Bigg] \nonumber\\
&= -\frac{1}{2} \sum_{i=1}^d \Big[ \ln{\sigma_i^2}  - \ln\gamma^2 + 1 - \frac{\sigma_i^2}{\gamma^2} - \frac{\mu_i^2}{\gamma^2} \Big] \nonumber\\
&= D_{KL}(P_{Z|X}\|Q_{Z}).
\end{align}
\subsection{Derivation of $h_{\alpha}\left(P_{Z \mid X} \| Q_{Z}\right)$ for Gaussians}\label{derivation-cross-ent}
Here we calculate the (differential) R\'enyi cross-entropy term of our cost function when $P_{Z|X}$ and $Q_Z$ are multivariate Gaussian distributions with $P_{Z|X}$ having a diagonal covariance structure while $Q_Z$ being a spherical Gaussian. Again we let 
\begin{align*}
P_{Z|X}&= \mathcal{N}\left(Z|\mu_{\text enc}(X),\text{diag}(\sigma_{\text enc}^2(X)) \right) \\ 
Q_Z &= \mathcal{N}(Z|\underbar{0},\gamma^2I_d),
\end{align*}
where $\mu_{\text enc}$ and $\sigma_{\text enc}^2$ are $d$-dimensional mean and variance vectors (that depend on $X$), $\text{diag}(\sigma_{\text enc}^2(X))$ is a $d\times d$ diagonal matrix with the entries of $\sigma_{\text enc}^2$ on the diagonal, $\underbar{0}$ is the all-zero vector of size $d$, $\gamma$ is a positive scalar, and $I_d$ is the $d$-dimensional identity matrix. For simplicility, we write $\mu_{\text enc}(X)$ as $\mu_{\text enc}$ and $\text{diag}(\sigma_{\text enc}^2(X))$ as $\sigma^2_{\text enc}I_d$.

Starting with the closed-form expression of the R\'enyi cross-entropy derived in \cite{thierrin2022}, we have
\begin{align}
&h_{\alpha}\left(P_{Z|X};Q_Z\right)
\nonumber  \\
&= \frac{1}{2-2 \alpha}\left(-\ln |A|\left|\sigma_{enc}^2I_d\right| 
+(1-\alpha) \ln (2 \pi)^d\left|\gamma^2I_d\right|-G\right)
\end{align}
where 
$$A=(\sigma_{enc}^2I_d)^{-1}+(\alpha-1) (\gamma^2I_d)^{-1}$$ 
is positive definite, and
\begin{align*}
G &= \nonumber \mu_{\text enc}'(\sigma_{enc}^2I_d)^{-1}\mu_{\text enc} 
\\
&\quad -\mu_{\text enc}'(\sigma_{enc}^2I_d)^{-1}A^{-1}(\sigma_{enc}^2I_d)^{-1}\mu_{\text enc},  
\end{align*}
with $\mu_{\text enc}'$ denoting the transpose of $\mu_{\text enc}$.
Then 
\begin{align}
-&\ln |A|\left|\sigma_{enc}^2I_d\right| \nonumber \\
=&-\ln \left|\begin{array}{ccc}
\frac{\gamma^2 +\sigma_1^2(\alpha-1)}{\sigma_1^2\gamma^2} & \cdots & 0 \\
\vdots & \ddots & \vdots \\
0 & \cdots & \frac{\gamma^2 +\sigma_d^2(\alpha-1)}{\sigma_d^2\gamma^2}
\end{array}\right| \left|\begin{array}{ccc}
\sigma_1^2 & \cdots & 0 \\
\vdots & \ddots & \vdots \\
0 & \cdots & \sigma_d^2
\end{array}\right| \nonumber\\
=&-\sum_{i=1}^d \ln\left(\frac{\gamma^2+\sigma_i^2(\alpha-1)}{\gamma^2}\right)
\end{align}
and 
\begin{equation}
-G = -\sum_{i=1}^d\frac{\mu_i^2}{\sigma_i^2}+\sum_{i=1}^d \frac{\mu_i^2}{\sigma_i^4}\frac{\sigma_i^2\gamma^2}{\gamma^2 +\sigma_i^2(\alpha-1)},
\end{equation}
where $\mu_i$ and $\sigma_i$ are the $i$th components of $\mu_{\text enc}$ and $\sigma_{\text enc}$,
yielding that
\begin{align}
h_{\alpha}\left(P_{Z|X};Q_Z\right) 
&= -\frac{1}{2-2\alpha}\sum_{i=1}^d \ln\left(\frac{\gamma^2+\sigma_i^2(\alpha-1)}{\gamma^2}\right)  \nonumber \\
&+ \frac{1}{2-2\alpha}(1-\alpha)\gamma^{2d}\ln(2\pi)^d \nonumber \\ 
-& \frac{1}{2-2\alpha} \sum_{i=1}^d \frac{\mu_i^2}{\sigma_i^2} -\frac{\mu_i^2}{\sigma_i^4}\frac{\sigma_i^2\gamma^2}{\gamma^2 +\sigma_i^2(\alpha-1)}. 
\label{eq:renyiCrossEntropy}
\end{align}

Since we require the matrix $A$
to be positive definite\footnote{\label{footnote4} As stated in Remark~1, we also require $\prod_{i=1}^d \sigma_i^2 \geq \frac{1}{\left(2 \pi e\right)^d}$ to ensure that 
$h(Z \mid X=x)=\frac{1}{2} \log \big((2 \pi e)^d \prod_{i=1}^d \sigma_i^2\big)
\geq 0.$ This is satisfied in our simulations as $d$ is large (we use $d=64$). \label{add-cond}}, the above \Renyi cross-entropy expression is valid for 
$$\gamma^2+\sigma_i^2(\alpha-1) > 0, \qquad i=1,\ldots,d, $$
or equivalently (recalling that $\alpha>0$, $\alpha\neq 1$) for
\begin{align} \nonumber
0 < \alpha < 1,  \ \sigma_i^2 < \frac{\gamma^2}{1-\alpha},& \qquad   \\
\alpha > 1, \ \sigma_i^2 >0,& \qquad i=1,\ldots,d. \label{eq:condition2}
\end{align}
\section{Experiments}
We conduct experiments on three different image datasets: CelebA, FairFace, and EyePACS, and on the tabular Adult and COMPAS datasets.\footnote{The code of our RFIB algorithm and all dataset partitions are available in full detail at the site: https://github.com/AGronowski/RFIB-Code.} In this section, we detail the implementation steps of our method, describe the metrics and the datasets we used, and explain how the experiments were performed and present results. We also present a uniform manifold approximation and projection (UMAP) clustering analysis that visualizes the effects of the hyper-parameter $\alpha.$

\subsection{Data}
\begin{figure}[htb]
\centering
\subfloat{\includegraphics[width=.14\textwidth]{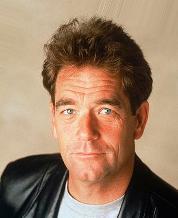}%
}
\subfloat{\includegraphics[width=.14\textwidth]{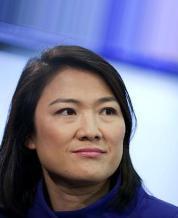}%
}
\subfloat{\includegraphics[width=.14\textwidth]{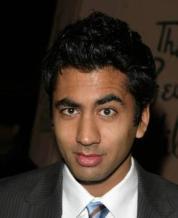}%
}
\vspace{-0.1in}
\subfloat{\includegraphics[width=.14\textwidth]{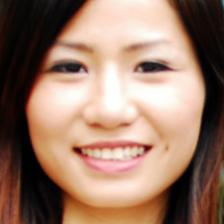}%
}
\subfloat{\includegraphics[width=.14\textwidth]{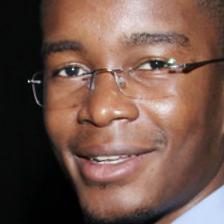}%
}
\subfloat{\includegraphics[width=.14\textwidth]{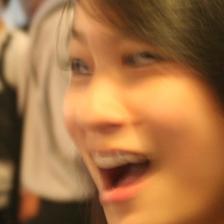}%
}
\vspace{-0.1in}
\subfloat{\includegraphics[width=.14\textwidth]{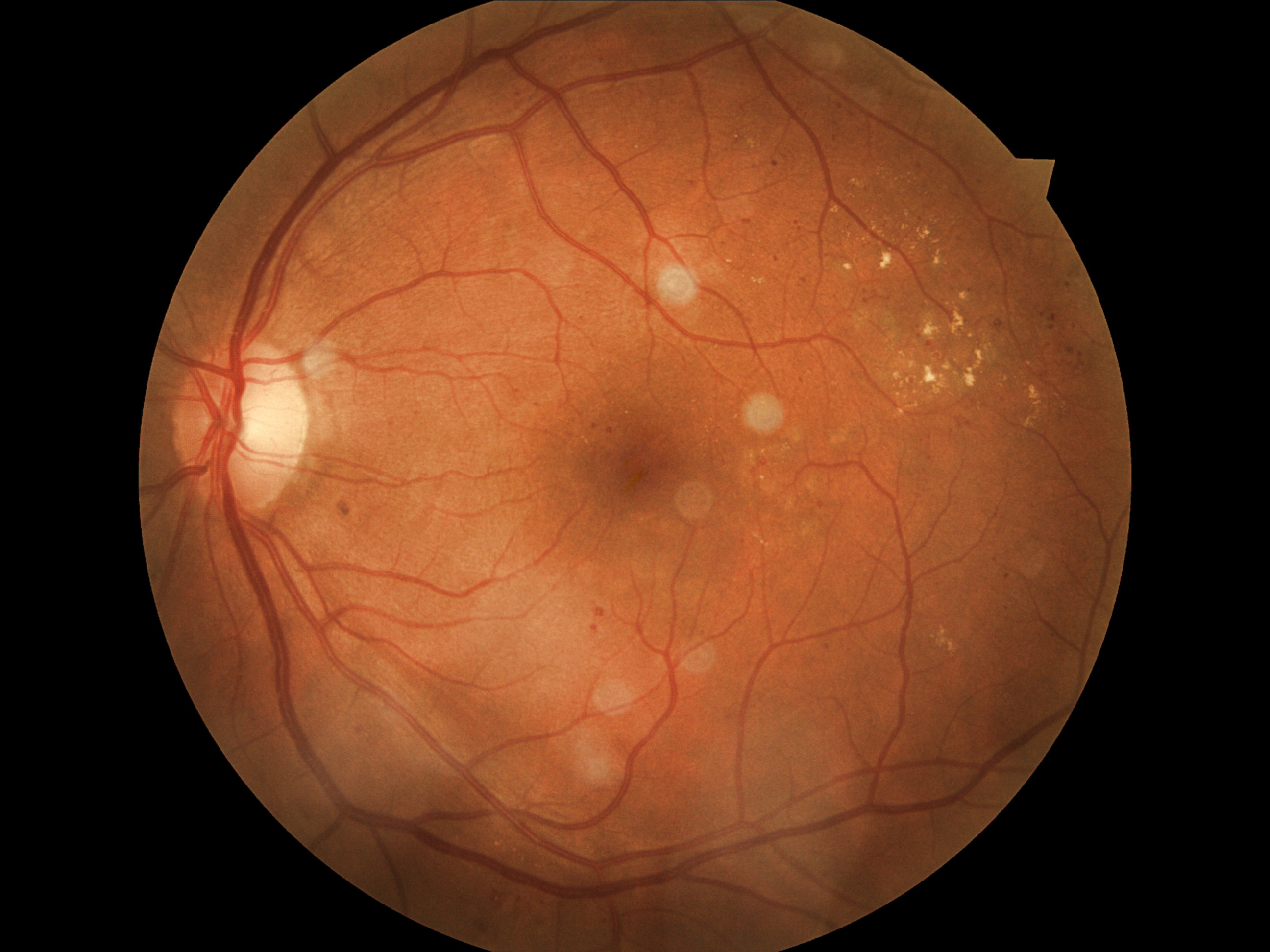}%
}
\subfloat{\includegraphics[width=.14\textwidth]{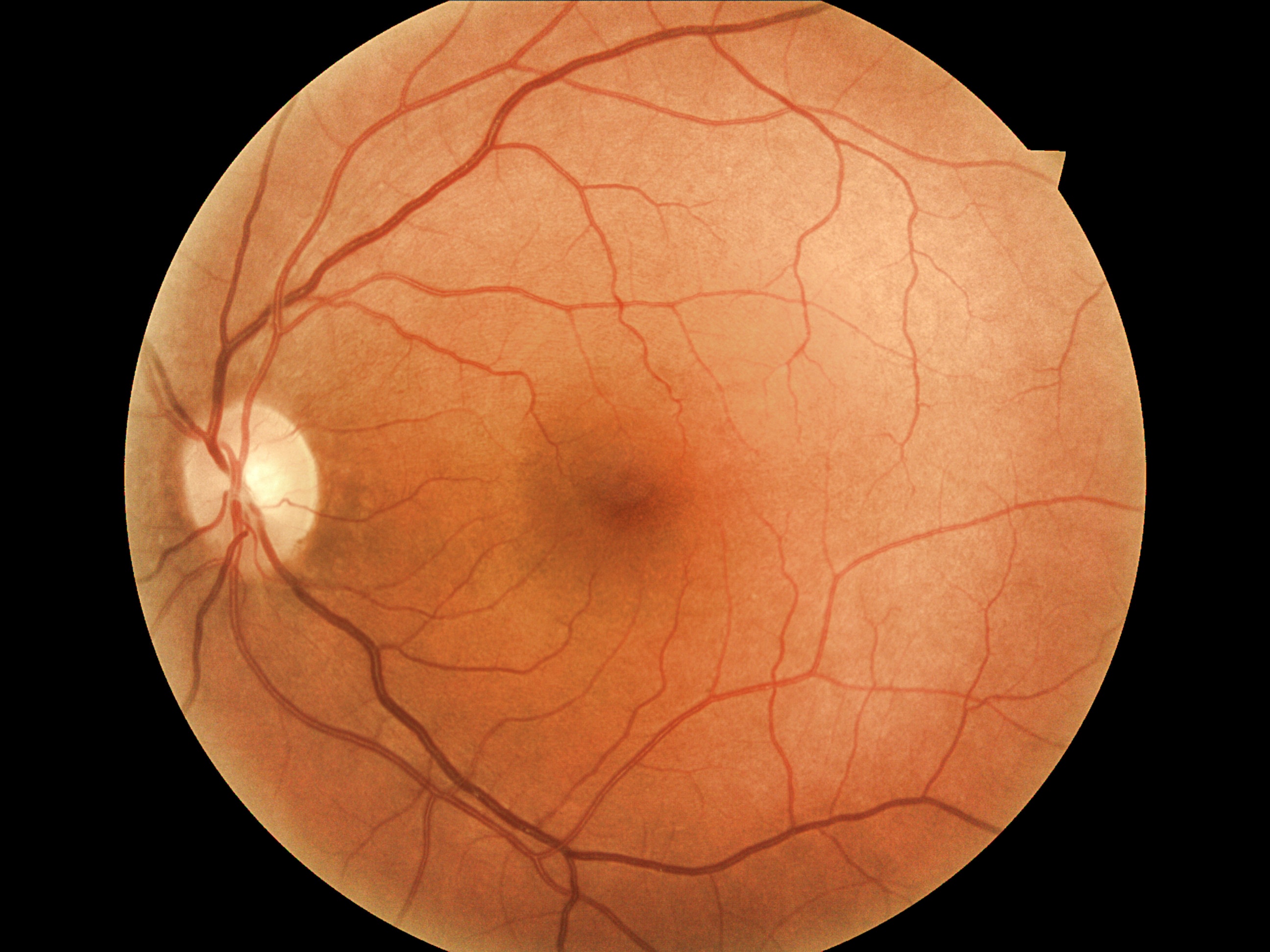}%
}
\subfloat{\includegraphics[width=.14\textwidth]{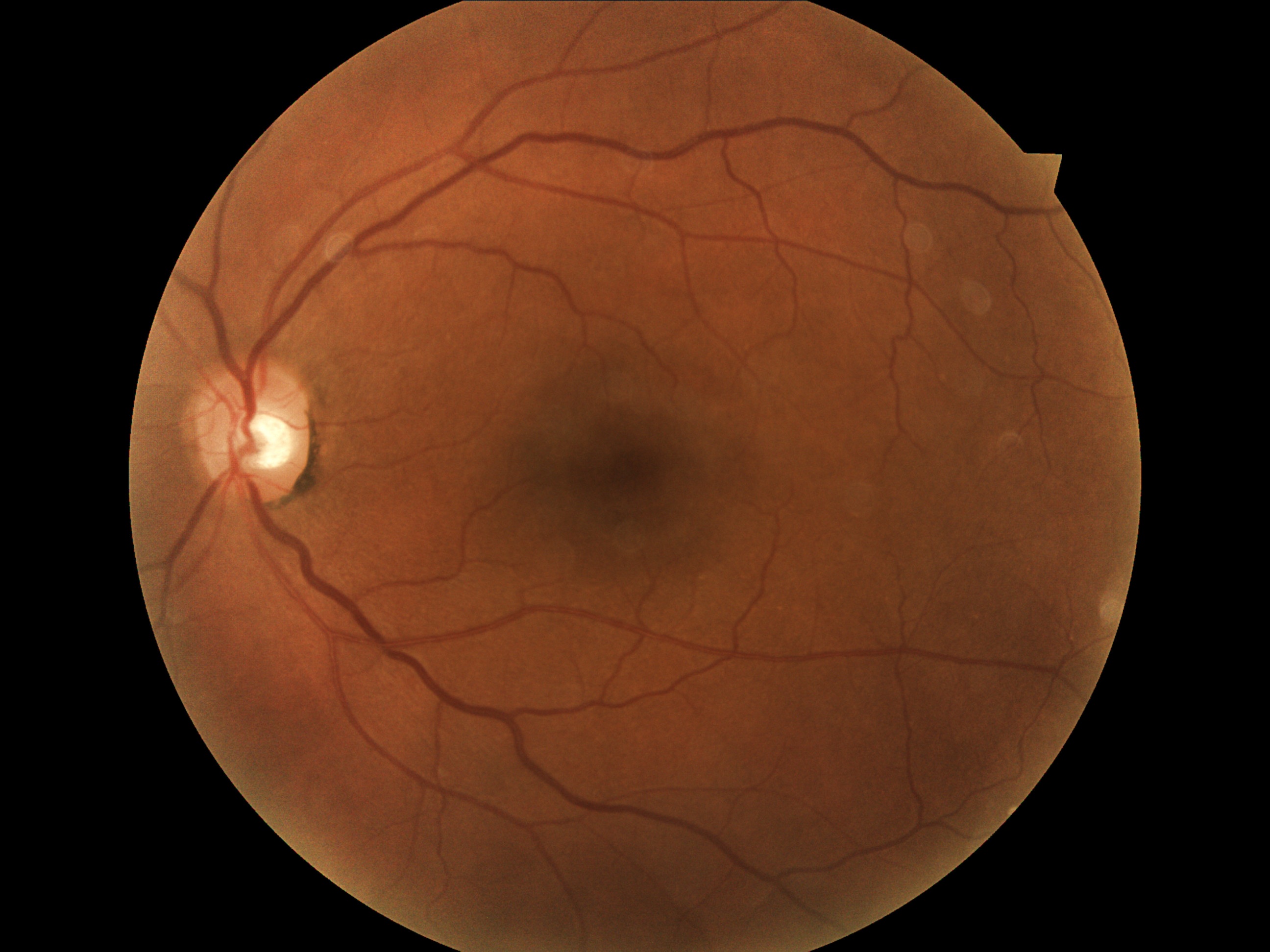}%
}
\caption{Examples of images taken respectively from  CelebA (top row), FairFace (middle row), and EyePACS (bottom row) datasets. For the EyePACS images, for the left image $(Y,S) = (1,0)$, for the middle $(Y,S) = (0,0)$, and for the right $(Y,S) = (0,1)$. }
\label{data}
\end{figure}

We next describe the image and tabular datasets (with the target $Y$ taken to be binary). Samples from the image datasets are shown in Fig.~\ref{data}.

\subsubsection{CelebA} The CelebA dataset \cite{liu2015faceattributes} contains 202,599 celebrity faces that each have 40 binary attributes. We use age as our prediction target $Y$, with $Y = 1$ indicating the person is old (according to the perception of the dataset's annotators), and are interested in skin tone as the sensitive attribute $S.$ As this is not included in the dataset we instead use the Individual Topology Angle (ITA) \cite{wilkes2015fitzpatrick} as a proxy, which was found to correlate with the Melanin Index, frequently used in dermatology to classify human skin on the Fitzpatrick scale.
As in \cite{merler2019diversity,paul2020tara}, we compute ITA via 
\begin{equation}
\mathrm{ITA}=\frac{180}{\pi} \arctan \left(\frac{L-50}{b}\right),
\end{equation}
where $L$ is luminescence and $b$ is yellowness in CIE-Lab space. We then binarize ITA where an ITA of $\leq 28$ is taken to mean dark skin, matching category thresholds used in \cite{kinyanjui2019estimating,paul2020tara}.

\subsubsection{FairFace} The FairFace dataset \cite{karkkainenfairface} consists of 108,501 face images labeled with race, gender, and age. We consider gender as the target $Y$ and race as sensitive information $S.$ The dataset contains several different categories for race. We include experimental results where we binarize the race labels to Black and non-Black. We also include experiments where $S$ is a categorical variable with three race categories: Black, Latino, and Other, with the Other group consisting of a mix of Indian, East Asian, Southeast Asian, Middle Eastern, and White races.

\subsubsection{EyePACS} The EyePACS dataset \cite{eyepacs} is sourced from the Kaggle Diabetic Retinopathy challenge. It consists of 88,692 retinal fundus images of individuals potentially suffering from diabetic retinopathy (DR), an eye disease associated with diabetes that is one of the leading causes of visual impairment worldwide. The dataset contains 5 categories of images based on the severity of the disease, with 0 being completely healthy and 4 being the most severe form of the disease. Similar to \cite{paul2020tara}, we binarize this label into our prediction target $Y,$ with $Y = 1$ corresponding to categories 2-4,  considered a positive, referable case for DR and $Y = 0$ corresponding to categories 0-1, considered healthy.

In our experiments we consider skin tone as the sensitive attribute. As with the CelebA dataset, we use ITA as a proxy for skin tone, with $S$ a binary variable that denotes whether or not the ITA of the fundus is $\leq 19,$ denoting that the individual has dark skin, as done by ~\cite{paul2020tara}. This has the advantage of being significantly easier to determine compared to the expensive and time consuming process of having a clinician manually annotate images. We compare these results with additional experiments we run using a smaller partition of the dataset with the race labels manually added by a ophthalmologist as done in ~\cite{burlina2020addressing}, determined based on factors that might correlate with race such as the darkness of pigmentation in the fundus, thickness of blood vessels, and ratio of the optic cup size to optic disk size. 

\subsubsection{Adult}
The Adult dataset \cite{dua2017uci}, also known as the Census Income dataset, contains 48,842 records of census data from 1994. Each sample contains 15 binary features such as gender, age, or income. We take our prediction target $Y$ to be income, with $Y=1$ corresponding to an income higher than \$50,000. We take the sensitive attribute $S$ to be gender.

\subsubsection{COMPAS}
The ProPublica COMPAS dataset \cite{dieterich2016compas} contains 6,172 records of data on criminal defendants, including attributes such as the defendant's gender, age, and race, and their recidivism outcome within 2 years of their initial screening. We take our sensitive attribute $S$ to be race and our prediction target $Y$ to be the recidivism outcome.

\subsection{Implementation Details}\label{sec:implementationDetails}
For all experiments, we use an isotropic Gaussian distribution for the encoder with mean and variance learned by a neural network, $P_{Z|X} = \mathcal{N}(Z|\mu_\text{enc},\sigma^2_\text{enc}I_d)$, using the same notation as described in Sections~\ref{derivation-div} and~\ref{derivation-cross-ent}.
Leveraging the reparameterization trick, we compute our representation  $Z \mid X$ as $ Z \mid X = \mu_\text{enc} + \sigma_\text{enc} \odot E$, where $\odot$ is the element-wise (Hadamard) product and $E\sim \mathcal{N}(0,I_d).$ 

We model the approximation of the representation's marginal as a $d$-dimensional spherical Gaussian, $Q_Z = \mathcal{N}(Z|\underbar{0},\gamma^2I_d).$
We calculate the R\'{e}nyi divergence in \eqref{final-loss} between the multivariate Gaussians $P_{Z|X}$ and $Q_Z$ using \eqref{eq:renyiDivergence} and calculate the \Renyi cross-entropy using (\ref{eq:renyiCrossEntropy}). For all experiments, we use a value of $\gamma = 1$. We add an additional sigmoid function to the encoder, limiting outputs such that all values of $\sigma_\text{enc}^2$ are $< 1$, ensuring the conditions in (\ref{eq:condition}), (\ref{eq:condition2}) and Footnote~\ref{add-cond} are satisfied. 

For binary target $Y$, we model $Q_{Y|Z}$ and $Q_{Y|Z,S}$ with Bernoulli distributions, $Q_{Y|Z} =\text{Bernoulli}(Y;f(Z))$ and $Q_{Y|Z,S} =\text{Bernoulli}(Y;g(Z,S))$, where\footnote{The notation $P_U=\text{Bernoulli}(U;p)$ means that $U$ is a Bernoulli random variable with parameter $p$ (i.e., $P(U=1)=p$).} $f$ and $g$ are auxiliary fully connected networks with a sigmoid function at the output, ensuring that $0 < f(Z) < 1$ and $0 < g(Z,S) < 1$. As specified in our description of the datasets above, we only use binary values for $Y$ in our experiments. However, our method can also accommodate
categorical (non-binary) values for $Y$, i.e., $\mathcal{Y} = \{0,\dots,k-1\}$ for integer $k > 2$. In this case, $Q_{Y|Z}$ and $Q_{Y|Z,S}$ can be modeled with Multinoulli distributions, $Q_{Y|Z} =\text{Multinoulli}\left(Y;f(Z)=(p_0,\dots,p_{k-1})\right)$ and $Q_{Y|Z,S} =\text{Multinoulli}\left(Y;g(Z,S)=(p_0,\dots,p_{k-1})\right)$, where\footnote{The notation $P_U=\text{Multinoulli}(U;(p_0,\dots,p_{k-1}))$ means that $U$ is a Multinoulli random variable with parameters $p_0,\dots,p_{k-1}$ (i.e., $P(U=i)=p_i$ for $i \in \{0,\dots,k-1$\}).} $f$ and $g$ are auxiliary fully connected networks with a softmax function at the output, ensuring that $\sum_{i=0}^{k-1}p_i = 1$ and $0 < p_i < 1$ for each $i \in \{0,\dots,k-1\}.$

While most prior work on fairness (e.g., \cite{zemel2013learning,hardt2016, zhao2019conditional,paul2020tara,skoglund2020} among others) consider only binary sensitive attributes, we also use our method with categorical sensitive attributes. When $S$ is binary, we use $S$ directly with (\ref{final-loss}), while when $S$ is categorical, i.e., $\mathcal{S} = \{0,\dots,n-1\}$ for integer $n > 2,$ we encode it as a one-hot encoding following standard practice in the literature~\cite{mougan2022fairness}. We then replace $S$ with its one-hot encoding in our final cost function; i.e., $S=(S_0, \dots, S_{n-1})=:S_0^{n-1}$, where each $S_j$, $j \in \{0,\dots,n-1\},$ is a binary random variable and component of the one-hot encoding vector of length $n$, leading to an RFIB final cost function of
\begin{align}\label{final-loss2}
    &J_\text{RFIB} 
    = \frac{1}{N}\sum_{i=1}^{N} \Big[ F_{\alpha}(P_{Z|X=x_i},Q_Z) \nonumber \\
    &- \beta_1 \mathbb{E}_E\left[\log \left(Q_{Y|Z}\left(y_i|f(x_i,E)\right)  \right)  \right] 
    \\&- \beta_2 
     \mathbb{E}_E\left[\log \left(Q_{Y|S_0^{n-1},Z}\left(y_i|s_{0,i},\dots, s_{n-1,i},f(x_i,E)\right)  \right)  \right]\Big]. \nonumber
\end{align}

The encoder network $P_{Z|X}$ is a ResNet50~\cite{He_2016_CVPR} classifier pretrained on ImageNet with the final linear layer replaced by two randomly initialized layers with output dimension $d$ equal to the dimension of the representation. The two decoder networks $f$ and $g$ are each two fully connected layers with 100 units and a sigmoid layer.

\begin{figure}[htb]
    \centering
    \includegraphics[width=.48\textwidth]{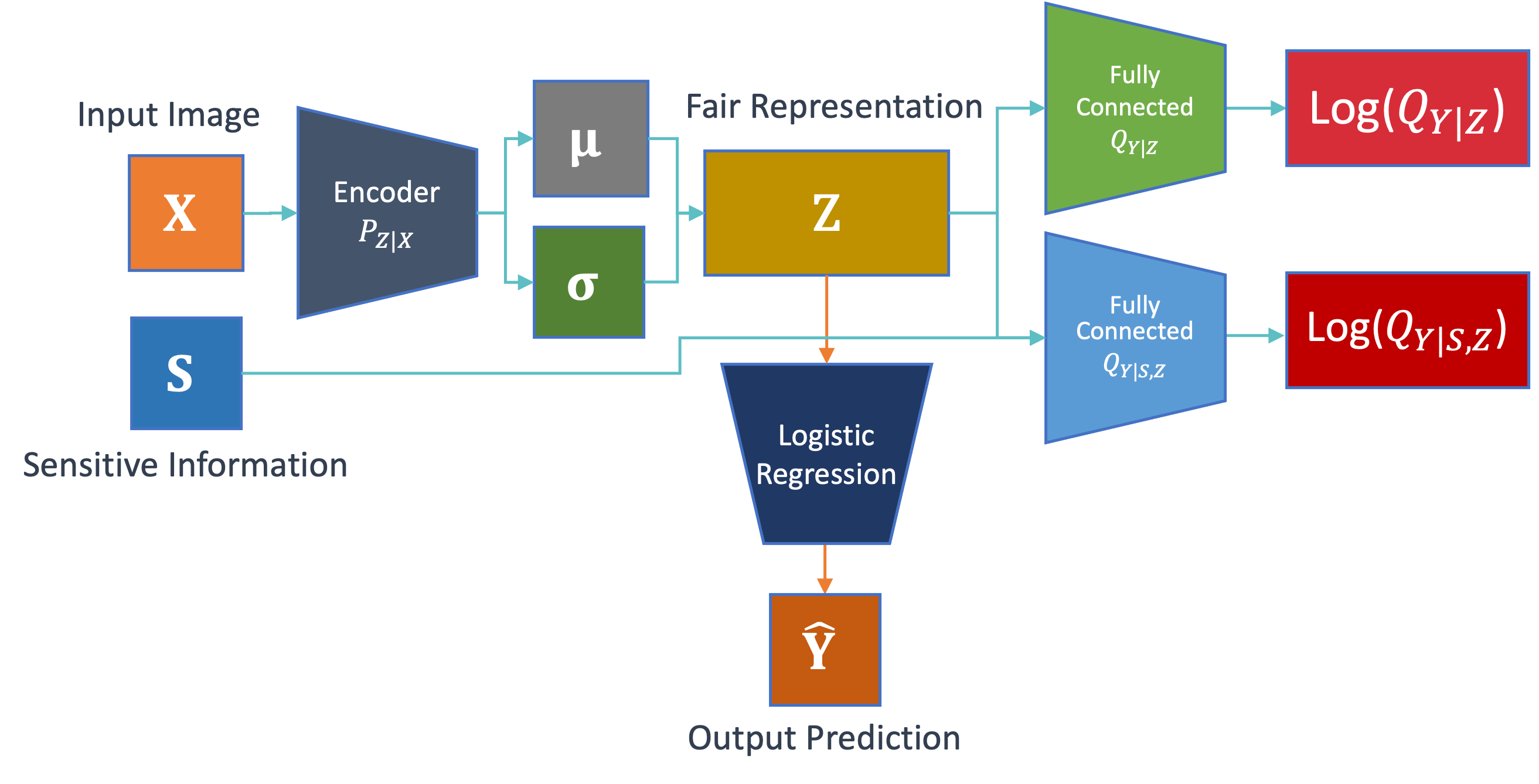}
    \caption{Architecture of the model is depicted. The input image $X$ is given to the encoder $P_{Z|X}$ that generates the mean $\mu$ and standard deviation $\sigma$ for the distribution of the fair representation $Z$, $\mathcal{N}(Z|\mu_\text{enc},\sigma^2_\text{enc}I_d)$. Then, $Z$ is  given to two fully connected networks, one of which with additional access to the sensitive information $S$. After training, $Z$ can be used as input to other existing architectures for fair prediction. For our experiments, we use logistic regression to predict $Y$ from $Z$.}
    \label{fig:diagram}
\end{figure}

After creating the representation $Z$, we use a logistic regression classifier with default settings 
to predict $Y$ from $Z$. We
evaluate accuracy and fairness on these predictions. Fig. \ref{fig:diagram} shows the architecture of our model.

We preprocess images by taking a 128 by 128 pixel center crop of the CelebA images, a 256 pixel by 256 pixel center crop of the EyePACS images, and use the full 224 pixel by 224 pixel images for FairFace. For the tabular Adult and COMPAS datasets, we normalize the input data $X$ to have zero mean and unit variance. We split the data into a training set and test set based on partitions described in detail in Section~\ref{sec:experimentDetails}. We then further split the training set, randomly choosing 10 \% of its data for validation, and use this validation set for early stopping. We train for up to 20 epochs, with early stopping triggering when there is no decrease in validation loss for 5 epochs, using a \texttt{min\_delta} value of 0.

For all experiments, we train using PyTorch on a NVIDIA GP100 GPU. We use $d = 64$ as the dimension of our representation $Z$, a batch size of 64, and the Adam optimizer with a learning rate of 0.001. For the experiments on image data with binary sensitive attribute $S$, we use the \Renyi divergence of order $\alpha$ in \eqref{final-loss}, by varying $\alpha$ linearly from 0 to 1, with $\alpha=0$ signifying that
\[
D_0(P_{Z|X}\|Q_Z) = -\log Q_Z(z: P_{Z|X}(z)>0)=0
\]
and 
with $\alpha = 1$ corresponding to $D_{\alpha}(P_{Z|X} \| Q_Z)$ being given by the KL divergence $D_{KL}(P_{Z|X} \| Q_Z)$; see~\eqref{extended-orders}. We also perform some additional experiments with $\alpha$ values $>1$, typically obtaining similar performance behavior as for $\alpha$ values $<1$.

For the experiments on tabular data with binary $S$ and on the image data with categorical $S$ (FairFace), we use the \Renyi cross-entropy in~\eqref{final-loss} and~\eqref{final-loss2}, respectively, with its order denoted by $\alphat$ (to differentiate it from the \Renyi divergence order $\alpha$), which we vary from 0 to~1.

\subsection{Metrics}
We use the following metrics to evaluate how well the model performs:

\subsubsection{Measure of Utility}
We use the overall classification accuracy (later denoted $\acc$):
    \begin{align}
    \acc = P(\hat{Y}= Y).
    \end{align}
\subsubsection{Measures of Fairness} We measure this in multiple ways: 
\begin{enumerate}[\ \ \ \ a)]
    \item using the gap in accuracy (denoted $\accgap$) between favored and protected subpopulations;
    \item reporting the minimum accuracy across subpopulations (denoted as $\accmin$), which is based on the Rawlsian principle of achieving fairness by maximizing $\accmin$ \cite{rawls2001justice};
    \item measuring the adherence to demographic parity via its gap $\dpgap$, following~\cite{skoglund2020};
    \item measuring the adherence using equalized odds via its gap $\eqoddsgap$.
\end{enumerate} 

These metrics are defined for only binary sensitive variables in prior work \cite{zemel2013learning, hardt2016, zhao2019conditional}, but here we generalize them to also handle non-binary (categorical) sensitive variables where 
$$\mathcal{S} =  \{0,1, \dots, n -1\}, \quad n \geq 2.$$
We define them as follows:
\begin{align}
&\accgap = \nonumber \\    
&\frac{1}{n(n-1)} \sum_{i=0}^{n-1}\sum_{\substack{j \neq i \\ j=0}}^{n-1}|P(\hat{Y}=Y |S=i)-P(\hat{Y}=Y |S=j)|, 
\end{align}
\begin{flalign}
&\accmin = \min_{s \in \mathcal{S}} P(\hat{Y} = Y | S = s), &&
\end{flalign}
\begin{flalign}
&\dpgap = \nonumber \\
&\frac{1}{n(n-1)} \sum_{i=0}^{n-1} \sum_{\substack{j \neq i \\
j=0}}^{n-1} | P(\hat{Y}=1 | S=i)-P(\hat{Y}=1| S=j) |,&&
\end{flalign}
\begin{flalign}
&\eqoddsgap = \nonumber \\
\begin{split}
 \max_{y \in \{0,1\}} \frac{1}{n(n-1)}\sum_{i=0}^{n-1} \sum_{\substack{j \neq i \\
j=0}}^{n-1} | P(\hat{Y}=1 | S=i, Y = y) \\ 
-P(\hat{Y}=1 | S=j, Y = y) |
\end{split} &&
\end{flalign}
where the last two metrics assume a binary target $Y$.
Note that for the binary sensitive attributes ($n=2$), these definitions recover the standard definitions from the literature.

\subsubsection{Joint Utility-Fairness Measure}  Echoing and comparing with the work in \cite{paul2020tara}, we use a single metric that jointly captures utility and fairness, the Conjunctive Accuracy Improvement ($\text{CAI}_{\lambda}$) measure: 
\begin{equation}
\text{CAI}_{\lambda} = \lambda (\accgap^\mathtt{b} - \accgap^\mathtt{d}) + (1 - \lambda) (\acc^\mathtt{d} - \acc^\mathtt{b})
\end{equation} 
where $ 0 \leq \lambda \leq 1 $, and $\acc^\mathtt{b}$ and $\acc^\mathtt{d}$ are the accuracy for baseline and debiased algorithms, respectively, while $\accgap^\mathtt{b}$ and $\accgap^\mathtt{d}$ are gap in accuracy for the baseline and debiased algorithms. In practice, one can use $\lambda=0.5$ for an equal balance between utility and fairness or a higher value such as $\lambda=0.75$ to emphasize fairness.

\subsection{Experiment Details and Results}\label{sec:experimentDetails}

We perform experiments on the EyePACS, CelebA, and FairFace datasets, considering the challenging case of severe data imbalance where training data is completely missing for one protected subgroup (e.g., diseased dark skin individuals for EyePACS). We also perform additional experiments on the Adult and COMPAS datasets.

\subsubsection{Hyper-parameter Tuning}
For all datasets and experiments, we conduct a hyper-parameter sweep.  We use various combinations of $\alpha$ (for \Renyi divergence) and $\tilde{\alpha}$ (for \Renyi cross-entropy) varied linearly from 0 to 1 (in addition to $\alpha$ values $>1$)
and $\beta_1$ and $\beta_2$ varied linearly from 1 to 50 (image datasets) or 1 to 100 (tabular datasets). Results are presented in Tables~\ref{tab:eyepacs}-\ref{tab:fairfacecat}, 
and are described in detail later in this section.


As mentioned in Section~\ref{compute-bounds}, the RFIB parameter pairs $(\alpha=1,\beta_1=0)$ and $(\alpha=1,\beta_2=0)$ correspond to the CFB and IB systems, respectively. We therefore compare RFIB with those two methods which are based on the KL divergence and only have a single $\beta$ hyper-parameter.
We do separate comparisons of CFB,
IB with RFIB, picking a commonly used value that performs well for CFB, IB and setting RFIB’s corresponding hyper-parameter to the
same value. We then use grid search to tune the two additional hyper-parameters that our method introduces; to compare with IB we fix a value of $\beta_1$ and tune $\alpha$ and $\beta_2,$ while to compare with CFB we fix a value of $\beta_2$ and then tune $\alpha$ and $\beta_1.$

We implement the IB and CFB methods ourselves and also compare RFIB with two methods used by \cite{paul2020tara}: adversarial independence, referred to as adversarial debiasing (AD), that minimizes conditional dependence of predictions on sensitive attributes with an adversarial two player game and intelligent augmentation (IA) that generates synthetic data for underrepresented populations and performs data augmentation to train a less biased model. To compare with AD and IA, we take results from \cite{paul2020tara} and report their original $\text{CAI}$ scores calculated with respect to their baseline, while for IB and CFB we implement the methods ourselves and calculate $\text{CAI}$ scores with respect to results from our own baseline, a ResNet50 network~\cite{He_2016_CVPR}. 

We recapitulate most acronyms used in Table \ref{tab:acronyms}.

\begin{table}[tbh]
\centering
\caption{List of acronyms used.}
\begin{tabular}{ll}
\textbf{Acronym} & \textbf{Stands for  }                      \\
\toprule
AD      & Adversarial Debiasing             \\
$\text{CAI}$    & Conjunctive Accuracy Improvement  \\
CFB     & Conditional Fairness Bottleneck   \\
DI & Disparate Impact \\
DR &    Diabetic Retinopathy \\
EO & Equalized Odds \\
IA      & Intelligent Augmentation          \\
IB      & Information Bottleneck            \\
ITA     & Individual Topology Angle         \\
KL      & Kullback-Leibler\\
RFIB    & \Renyi Fair Information Bottleneck \\
UMAP    & Uniform Manifold Approximation and Projection
\end{tabular}
\label{tab:acronyms}
\end{table}

\subsubsection{EyePACS Results}
We predict $Y=$ DR status while using $S=$ ITA as the sensitive attribute. We consider a scenario where training data is completely missing for the subgroup of $(Y,S) = (1,1)$, individuals referable for DR who have dark skin. The goal of our method is for predictions on the missing subgroup to be just as accurate as on the group with adequate training data, which is a problem of both fairness and also \textit{domain adaptation}, achieving high performance on a group not present in the original dataset. This scenario matches an important real world problem where data for a minority subgroup such as dark skinned individuals is lacking.

We create a training partition containing both images referable and non-referable for DR of light skin individuals but only non-referable images of dark skin individuals. We use the same partition as in \cite{paul2020tara} to compare with their method, using a training set that consists of 10,346 light skin images referable for diabetic retinopathy (DR = 1, ITA  = 0), 5,173 non-referable light skin images (DR = 0, ITA = 0), and 5,173 non-referable dark skin images (DR = 0, ITA = 1). Then for a fair assessment of our method's performance we evaluate on a balanced test set with an equal number of positive and negative examples for both dark and light skin individuals, with the set containing 2,400 images equally balanced across DR and ITA. 

As shown in Table \ref{tab:eyepacs}, our method outperforms all other methods, showing  improvements in accuracy and fairness across all metrics. We show a result with a value of $\alpha > 1$ in the bottom row, showing that it is still possible to achieve promising performance with higher $\alpha$ values, although slightly better results for most metrics were achieved with values of $\alpha < 1$. 
Usual caution should be exercised in interpretations since -- despite our aligning with data partitioning in~\cite{paul2020tara} -- other variations may exist with~\cite{paul2020tara,deepvariational,skoglund2020} due to non-determinism, parameter setting or other factors. Of interest is that we outperform the baseline in both fairness and accuracy. While somewhat unexpected, as we typically expect a trade-off between fairness and compression, this is potentially due to the compression term $I(Z;X)$ acting as a regularizer and improving generalization performance, as argued by \cite{deepvariational}.

We perform a second experiment where we use the same networks from before trained using $S = $ ITA but test on a balanced test set where $S = $ Ethnicity (defined as in \cite{paul2020tara}), with labels coming from a human clinician. The test set consists of 400 images equally balanced across DR and ethnicity. We show these experimental results in Table \ref{tab:eyepacs_ethnic}, where we outperform the other methods across most metrics, including the most important $\text{\text{CAI}}$ scores. As we achieve similar results using ITA and race labels from a human clinician, we show that ITA can successfully be used as an easily obtained alternative to manual label annotation, further supporting conclusions reached by \cite{paul2020tara}. These results also demonstrate the ability of our method to perform well in this type of protected factor domain adaptation problem where a different protected factor is used after initial training, which is important in settings where the actual protected factor is not revealed for privacy reasons.


\begin{table*}[tbh]
\caption{Results for debiasing methods on EyePACS predicting $Y$= DR Status, trained on partitioning with respect to $S=$ ITA, and evaluated on a test set balanced across DR status and ITA. For metrics with an $\uparrow$ higher is better whereas for $\downarrow$ lower is better. Subpopulation is the one that corresponds to the minimum accuracy, with (D) indicating dark skin and (L) light skin. Metrics are given as percentages. NA indicates that results are not available due to not being reported in the cited work.}
\centering
\begin{tabular}{l||c|c|c|c|c|c|c}
    \toprule
\multicolumn{1}{l||}{Methods}                               & \multicolumn{1}{l|}{$\acc \uparrow$} & \multicolumn{1}{l|}{$\accgap \downarrow $} & \multicolumn{1}{l|}{\begin{tabular}{@{}l@{}} $\accmin \uparrow$ \\  (subpop.) \end{tabular}} & \multicolumn{1}{l|}{$\text{\text{CAI}}_{0.5} \uparrow$} & \multicolumn{1}{l|}{$\text{\text{CAI}}_{0.75} \uparrow$} & \multicolumn{1}{l|}{$\dpgap \downarrow$} & \multicolumn{1}{l}{$\eqoddsgap \downarrow$} \\ \hline \hline
Baseline (results from~\cite{paul2020tara}) & 70.0                                & 3.5                                          & 68.3                                                   & -                                         & -                                          & NA                                         & NA                                            \\ \hline
AD ($\beta = 0.5$) (results from~\cite{paul2020tara})                                        & 76.12                               & 2.41                                         & 74.92 (L)                                          & 3.61                                      & 2.35                                       & NA                                         & NA                                            \\ \hline
IA (results from~\cite{paul2020tara})                                                        & 71.5                                & 1.5                                          & 70.16 (D)                                           & 1.75                                      & 1.875                                      & NA                                         & NA                                            \\ \hline
Baseline (ours)                                                   & 73.37                               & 8.08                                         & 69.33 (D)                                           & -                                         & -                                          & 28.25                                      & 36.33                                         \\ \hline
IB ($\beta_1$=30) \cite{deepvariational}                                         & 74.12                               & 2.08                                         & 73.08 (D)                                           & 3.37                                      & 4.69                                       & 18.58                                      & 20.67                                         \\ \hline
CFB ($\beta_2$=30) \cite{skoglund2020}                                       & 77.83                               & 1.66                                         & 77.0 (L)                                           & 5.44                                      & 5.93                                       & 10.83                                      & 12.5                                          \\ \hline
\rowcolor{LightCyan} \begin{tabular}{@{}l@{}} RFIB (ours)\\  ($\alpha = 0.8, \beta_1 = 36, \beta_2$ = 30) \end{tabular}          & 79.42                               & 0.5                              & \textbf{79.17 (L)  }                                        & \textbf{6.81   }                                 & \textbf{7.19     }                         & 16.17                                   & 16.67                                          \\ \hline
 
\rowcolor{LightCyan} \begin{tabular}{@{}l@{}} RFIB (ours)\\  ($\alpha = 0.3, \beta_1 = 30, \beta_2$ = 50) \end{tabular}                & \textbf{79.71    }              & 1.75                                         & 78.83 (L)                                 & 6.33                             & 6.33                                      & \textbf{9.75    }                          & \textbf{11.50 }                              \\ \hline

\rowcolor{LightCyan} \begin{tabular}{@{}l@{}} RFIB (ours)\\  ($\alpha = 1.8, \beta_1 = 30, \beta_2$ = 17) \end{tabular}     & 78.35                  & \textbf{0.25}                                         & 78.25 (L)                                 & 6.41                             & 7.12                                      & 15.58                             & 15.83                               \\ \hline

\bottomrule
\end{tabular}
\label{tab:eyepacs}
\end{table*}

\begin{table*}[tbh]
\caption{Performance results for debiasing methods on EyePACs predicting $Y$= DR Status, trained on partitioning with respect to $S=$ ITA, and evaluated on a test set balanced across DR status and ethnicity (defined as in [35]). 
For metrics with an $\uparrow$ higher is better whereas for $\downarrow$ lower is better. Subpopulation is the one that corresponds to the minimum accuracy, with (W) indicating white individuals and (B) Black individuals. Metrics are given as percentages.}
\label{tab:eyepacs_ethnic}
\centering
\begin{tabular}{l||c|c|c|c|c|c|c}
\toprule 
\multicolumn{1}{l||}{Methods}                             & \multicolumn{1}{l|}{$\acc \uparrow$} & \multicolumn{1}{l|}{$\accgap\downarrow$} & \multicolumn{1}{l|}{\begin{tabular}{@{}l@{}} $\accmin \uparrow$ \\  (subpop.) \end{tabular}} & \multicolumn{1}{l|}{$\text{\text{CAI}}_{0.5} \uparrow$} & \multicolumn{1}{l|}{$\text{\text{CAI}}_{0.75} \uparrow$} & \multicolumn{1}{l|}{$\dpgap\downarrow$} & \multicolumn{1}{l}{$\eqoddsgap\downarrow$} \\ \hline \hline
Baseline&76.00&13.00&69.50 (B)&-&-&3.00&16.00\\ \hline
IB ($\beta_2$=30) \cite{skoglund2020}&78.75&\textbf{3.50}&77.00 (B)&6.12&\textbf{7.81}&\textbf{0.50}&\textbf{4.00}\\ \hline
CFB ($\beta_2$=30) \cite{skoglund2020} &75.00&6.00&72.00 (B)&3.00&5.00&5.00&11.00\\ \hline
\rowcolor{LightCyan} \begin{tabular}{@{}l@{}} RFIB (ours)\\  ($\alpha = 0.8, \beta_1 = 36, \beta_2$ = 30) \end{tabular}&81.5&6.00&78.50 (B)&6.25&6.63&3.00&9.00\\ \hline
\rowcolor{LightCyan}\begin{tabular}{@{}l@{}} RFIB (ours)\\  ($\alpha = 0.3, \beta_1 = 30, \beta_2$ = 50) \end{tabular}&\textbf{81.75}&\text4.50&\textbf{79.50 (B})&\textbf{7.13}&\textbf{7.81}&11.50&16.00\\ \hline

\bottomrule
\end{tabular}
\end{table*}

\subsubsection{CelebA Results}
We predict age using ITA as the sensitive attribute. Again, we consider the domain adaptation/fairness problem where part of the data is completely missing for a protected subgroup. In this case, older light skin images are missing and we use a training set of 48,000 images, consisting of 24,000 older dark skin images, 12,000 younger light skin images, and 12,000 older dark skin images. We use a test set of 8,000 images equally balanced across age and ITA, using the same partition as \cite{paul2020tara} and varying hyper-parameters and comparing methods the same way as for EyePACS. Our results are shown in Table \ref{tab:celebarace}, showing our method performs the best across all metrics.


\begin{table*}[tbh]
    \caption{Results for debiasing methods on CelebA predicting $Y$= Age, trained on partitioning with respect to $S=$ ITA, and evaluated on a test set balanced across age and ITA. For metrics with an $\uparrow$ higher is better whereas for $\downarrow$ lower is better. Subpopulation is the one that corresponds to the minimum accuracy, with (D) indicating dark skin and (L) light skin. Metrics are given as percentages. NA indicates that results are not available due to not being reported in the cited work.}

    \centering

\begin{tabular}{l||c|c|c|c|c|c|c}
\toprule
\multicolumn{1}{l||}{Methods}                                & \multicolumn{1}{l|}{$\acc \uparrow$} & \multicolumn{1}{l|}{$\accgap \downarrow$} & \multicolumn{1}{l|}{\begin{tabular}{@{}c@{}} $\accmin \uparrow$ \\  (subpop.) \end{tabular}} & \multicolumn{1}{l|}{$\text{CAI}_{0.5} \uparrow$} & \multicolumn{1}{l|}{$\text{CAI}_{0.75} \uparrow$} & \multicolumn{1}{l|}{$\dpgap \downarrow$} & \multicolumn{1}{l}{$\eqoddsgap \downarrow$} \\ \hline \hline
Baseline (results from~\cite{paul2020tara})  & 74.4                                & 13.9                                        & 67.5                                                  & -                                         & -                                          & NA                                         & NA                                            \\ \hline
AD ($\beta = 0.5$) (results from~\cite{paul2020tara})  & 76.45                               & 9.6                                         &71.65 (D)                                          & 3.17                                      & 3.75                                       & NA                                         & NA                                            \\ \hline
IA (results from~\cite{paul2020tara})                & 75.29                               & 9.18                                        & 70.7 (D)                                           & 2.8                                       & 1.56                                       & NA                                         & NA                                            \\ \hline
\multicolumn{1}{l||}{Baseline (ours)}                        & 70.61                               & 16.57                                       & 62.32 (D)                                          & -                                         & -                                          & 43.82                                      & 60.4                                          \\ \hline
IB ($\beta_1$=30) \cite{deepvariational}                                         & 71.76                               & 14.87                                       & 64.32 (D)                                          & 1.42                                      & 4.69                                       & 36.62                                      & 51.5                                          \\ \hline
CFB  ($\beta_2$=30) \cite{skoglund2020}                                       & 71.77                               & 13.25                                       & 65.15 (D)                                          & 2.24                                      & 2.78                                       & 38.3                                       & 51.55                                         \\ \hline
\rowcolor{LightCyan} \begin{tabular}{@{}l@{}} RFIB (ours)\\  ($\alpha = 0.3, \beta_1 = 30, \beta_2$ = 43) \end{tabular}           & 75.04                               & 10.92                                        & 69.57 (D)                                          & 5.04                                      & 5.34                                       & 29.02                                     & 39.95                                        \\ \hline
\rowcolor{LightCyan} \begin{tabular}{@{}l@{}} RFIB (ours)\\  ($\alpha = 0.4, \beta_1 = 1, \beta_2$ = 30) \end{tabular}             & \textbf{76.92}                               & \textbf{5.05 }                                       & \textbf{74.4 (D) }                                         & \textbf{8.91   }                                   & \textbf{10.22}                                     & \textbf{2.7  }                                    & \textbf{7.75}                                         \\ \hline \bottomrule
\end{tabular}
\label{tab:celebarace}
\end{table*}

\subsubsection{Adult and COMPAS Results}
We predict income on the Adult dataset using gender as the sensitive attribute and predict recidivism outcome on the COMPAS dataset using race as the sensitive attribute. For Adult, we use the train/test partitions provided with the dataset, while for COMPAS we randomly split the data using a 70\%/30\% train/test split. We compare with two methods, Fair Classifier \cite{cho2020fair} and FR-Train \cite{roh2020fr}, using two fairness metrics defined in those works, Disparate Impact (DP) and Equalized Odds (EO):
\begin{equation}
\text{DI}=\min _{s \in \mathcal{S}} \min _{\hat{s} \neq s} \frac{P(\hat{Y}=1 | S=\hat{s})}{P(\tilde{Y}=1 | S=s)},
\end{equation}
\begin{equation}
\text{EO}=\min _{y \in \mathcal{Y}} \min _{s \in \mathcal{S}} \min _{\hat{s} \neq s} \frac{P(\hat{Y}=1 | S=\hat{s}, Y=y)}{P(\tilde{Y}=1 | S=s, Y=y)}.
\end{equation}
We show results in Tables~\ref{tab:adult} and~\ref{tab:compas} with various combinations of hyper-parameters including where we achieve the highest accuracy, highest DI, and highest EO. As expected, there is typically a trade-off between achieving high accuracy and
high performance on the fairness metrics (this is also confirmed in the recent work in~\cite{paul2023evaluating} based on the AD method). In most cases, higher $\beta$ values coupled with higher $\alphat$ values result in increased accuracy but worsened fairness (note that the RFIB compression term $I(Z;X)$ is characterized here by the \Renyi cross-entropy of order~$\alphat$, which is a non-increasing function of $\alphat$). 

Compared to Fair Classifier and FR-Train, we suffer a drop in accuracy to achieve a higher DI, but outperform these baselines with respect to EO (the most important fairness metric according to \cite{hardt2016}). On the Adult dataset, we achieve higher EO scores than both Fair Classifier and FR-Train while maintaining the same or very comparable accuracy. In cases where we do need to sacrifice accuracy to achieve fairness, we are still able to beat both baselines on at least one metric according to how we tune $\alphat$ and the two $\beta$ hyper-parameters. In those cases, which method is preferable will depend on the application and the metric that is most suitable for that application.




\begin{table*}[tbh]
\begin{minipage}{.5\linewidth}
\caption{Results for debiasing methods on Adult predicting $Y$= Income where $S$ is Race. The $\uparrow$ indicates that higher is better, and $\alphat$ indicates that results were obtained using \Renyi cross-entropy. NA indicates that results are not available due to not being reported in the cited work.}
\centering
\begin{tabular}{l||c|c|c}
\toprule 
\multicolumn{1}{l||}{Methods}                             & \multicolumn{1}{l|}{$\acc \uparrow$} & \multicolumn{1}{l|}{DI $\uparrow$} & \multicolumn{1}{l}{EO $\uparrow$} \\ \hline \hline
Fair Classifier (results from~\cite{cho2020fair})                                                                                                   & 83.7                                 & 0.942                          & NA                               \\ \hline
Fair Classifier  (results from~\cite{cho2020fair})                                  &\textbf{84.4 }                               & NA                            & 0.753                             \\ \hline
FR-Train (results from~\cite{roh2020fr})                                                                              & 82.4                             & 0.828                          & NA                             \\ \hline
FR-Train (results from~\cite{roh2020fr})                                                                              & 84.2                             & NA                          & 0.917                             \\ \hline
\rowcolor{LightCyan} \begin{tabular}{@{}l@{}} RFIB (ours)\\  ($\alphat = .3 , \beta_1 = 1 , \beta_2$ = 84 ) \end{tabular}                                 & 84.2                    & 0.420                & 0.970                      \\ \hline
\rowcolor{LightCyan} \begin{tabular}{@{}l@{}} RFIB (ours)\\  ($\alphat = .16 , \beta_1 = 1, \beta_2$ = 47 ) \end{tabular}                                 & 79.5                    & 0.492                & \textbf{0.999 }                    \\ \hline
\rowcolor{LightCyan} \begin{tabular}{@{}l@{}} RFIB (ours)\\  ($\alphat = .42, \beta_1 = 1, \beta_2$ = 0 ) \end{tabular}                                 & 76.2                    & \textbf{0.943 }               & 0.424                     \\ \hline

\bottomrule
\end{tabular}
\label{tab:adult}
\end{minipage}
\begin{minipage}{.5\linewidth} \centering 
\caption{Results for debiasing methods on COMPAS predicting $Y$= Recidivism Outcome where $S$ is Race. The $\uparrow$ indicates that higher is better, and $\alphat$ indicates that results were obtained using \Renyi cross-entropy. NA indicates that results are not available due to not being reported in the cited work.}
\centering
\begin{tabular}{l||c|c|c}
\toprule 
\multicolumn{1}{l||}{Methods}                             & \multicolumn{1}{l|}{$\acc \uparrow$} & \multicolumn{1}{l|}{DI $\uparrow$} & \multicolumn{1}{l}{EO $\uparrow$} \\ \hline \hline
Fair Classifier (results from~\cite{cho2020fair})                                                                                                   & 67.5                                 & 0.959                         & NA                               \\ \hline
Fair Classifier  (results from~\cite{cho2020fair})                                  &63.8                                & NA                            & 0.908                             \\ \hline
FR-Train (results from~\cite{roh2020fr})                                                                              & 67.6                             & 0.838                         & NA                             \\ \hline
FR-Train (results from~\cite{roh2020fr})                                                                              & 62.8                             & NA                          & 0.959                             \\ \hline
\rowcolor{LightCyan} \begin{tabular}{@{}l@{}} RFIB (ours)\\  ($\alphat = .6 , \beta_1 = 1 , \beta_2$ = 53 ) \end{tabular}                                 & \textbf{68.2 }                   & 0.571                & 0. 717                      \\ \hline

\rowcolor{LightCyan} \begin{tabular}{@{}l@{}} RFIB (ours)\\  ($ \alphat = .5 , \beta_1 = 1, \beta_2$ = 6 ) \end{tabular}                                 & 57.6                    & 0.783                & \textbf{0.979}                     \\ \hline

\rowcolor{LightCyan} \begin{tabular}{@{}l@{}} RFIB (ours)\\  ($\alphat = .3, \beta_1 = 1, \beta_2$ = 2 ) \end{tabular}                                 & 55.0                    & \textbf{0.999}                & 0.875                     \\ \hline

\bottomrule
\end{tabular}
\label{tab:compas}
\end{minipage} 
\end{table*}

\subsubsection{FairFace Results}
We predict $Y$ = Gender with $S$ = Race as the sensitive attribute, testing on a test set balanced across gender and race. As before, we exclude one population subgroup and remove Black females from the training data, matching the common real-world scenario where data is lacking for this group.  Unlike the previous experiments, here we can obtain race directly from the dataset rather than using ITA as a proxy for skin tone. We binarize the race labels into two groups, a smaller Black group and a larger non-Black group. We use a training set of 16,500 images, consisting of 5,500 male Black images, 5,500 male white images, and 5,500 female white images. We test on a test set of 3,000 images equally balanced across gender and race. Our results, given in Table~\ref{tab:fairface}, show that we outperform both the IB and CFB methods both on accuracy and on all fairness metrics. 

\begin{table*}[tbh]
\caption{Results for debiasing methods on FairFace predicting $Y$= Gender, trained on partitioning with respect to $S=$ Race, and evaluated on a test set balanced across Gender and ITA. For metrics with an $\uparrow$ higher is better whereas for $\downarrow$ lower is better. Subpopulation is the one that corresponds to the minimum accuracy, with (B) indicating Black. Metrics are given as percentages.}
    \centering

\begin{tabular}{l||c|c|c|c|c|c|c}
\toprule 
\multicolumn{1}{l||}{Methods}                             & \multicolumn{1}{l|}{$\acc \uparrow$} & \multicolumn{1}{l|}{$\accgap\downarrow$} & \multicolumn{1}{l|}{\begin{tabular}{@{}l@{}} $\accmin \uparrow$ \\  (subpop.) \end{tabular}} & \multicolumn{1}{l|}{$\text{CAI}_{0.5} \uparrow$} & \multicolumn{1}{l|}{$\text{CAI}_{0.75} \uparrow$} & \multicolumn{1}{l|}{$\dpgap\downarrow$} & \multicolumn{1}{l}{$\eqoddsgap\downarrow$} \\ \hline \hline
Baseline                                                 & 73.97                      & 16.47                            & 65.73 (B)                               & -                                & -                                 & 27.53                           & 44.0                               \\ \hline
IB  ($\beta_2$=30) \cite{deepvariational}                                     & 73.93                      & 14.93                            & 68.6 (B)                                & 0.75                             & 1.14                              & 30.0                            & 44.93                              \\ \hline
CFB ($\beta_2$=30) \cite{skoglund2020}                                      & 75.6                       & 14.0                             & 65.2 (B)                                & 2.05                             & 2.26                              & 26.27                           & 40.27                              \\ \hline
\rowcolor{LightCyan} \begin{tabular}{@{}l@{}} RFIB (ours)\\  ($\alpha = 0.2, \beta_1 = 30, \beta_2$ = 29) \end{tabular}     & \textbf{83.6}                     & 8.53                  & \textbf{79.33 (B)}                    & \textbf{8.78 }                  & 8.36                    & 19.47                & 28.0                      \\ \hline
\rowcolor{LightCyan} \begin{tabular}{@{}l@{}} RFIB (ours)\\  ($\alpha = 0.2, \beta_1 = 1, \beta_2$ = 30) \end{tabular}      & 82.07             & \textbf{7.33 }                           & 78.4 (B)                               & 8.62                             & \textbf{8.88  }                            & \textbf{19.07}                           & \textbf{26.4 }                              \\ \hline
\bottomrule
\end{tabular}

\label{tab:fairface}
\end{table*}

\subsubsection{FairFace Results (Categorical Variables)}

We extend our method to work with non-binary categorical attributes (as described in Section~\ref{sec:implementationDetails}). We predict $Y =$ Gender with $S =$ Race containing three race groups: Black, Hispanic, and Other, with the Other group containing a mix of White, East Asian, Southeast Asian, Middle Eastern, and Indian races. We use a training set of 17,500 images and test set of 7,500 images. Since most prior work on fairness is defined only for binary attributes, we compare RFIB with two non-fairness methods, a ResNet50 baseline and the IB method. We present results in Table~\ref{tab:fairfacecat}, showing that we are able to achieve high accuracy while also achieving improvement across all but one of the fairness metrics, most significantly for the $\dpgap$ and $\eqoddsgap$ metrics.
\begin{table*}[tbh]
\caption{Results for debiasing methods on FairFace predicting $Y$= Gender where $S$ is race, and $S$ is a categorical attribute with three race categories. For metrics with an $\uparrow$ higher is better whereas for $\downarrow$ lower is better, and $\alphat$ indicates that results were obtained using \Renyi cross-entropy. Subpopulation is the one that corresponds to the minimum accuracy, with (B) indicating Black. Metrics are given as percentages.}
    \centering
\begin{tabular}{l||c|c|c|c|c|c|c}
\toprule 
\multicolumn{1}{l||}{Methods}                             & \multicolumn{1}{l|}{$\acc \uparrow$} & \multicolumn{1}{l|}{$\accgap\downarrow$} & \multicolumn{1}{l|}{\begin{tabular}{@{}l@{}} $\accmin \uparrow$ \\  (subpop.) \end{tabular}} & \multicolumn{1}{l|}{$\text{CAI}_{0.5} \uparrow$} & \multicolumn{1}{l|}{$\text{CAI}_{0.75} \uparrow$} & \multicolumn{1}{l|}{$\dpgap\downarrow$} & \multicolumn{1}{l}{$\eqoddsgap\downarrow$} \\ \hline \hline
Baseline                                                 & 83.9                      & 6.77                            & 76.72 (B)                               & -                                & -& 8.17                           & 13.18                               \\ \hline
IB  ($\beta_2$=30) \cite{deepvariational}                                      & 79.12                      & \textbf{5.87 }                           & 73.38 (B)                                &   -1.94                          & -0.52                                                               & 7.07                            & 11.07                              \\ \hline
\rowcolor{LightCyan} \begin{tabular}{@{}l@{}} RFIB (ours)\\  ($\alphat = 0.2, \beta_1 = 30, \beta_2$ = 29) \end{tabular}     & \textbf{85.24}                     & 6.74                  & \textbf{78.01 (B)}                    & \textbf{0.68}                  &  \textbf{0.35}                      & \textbf{3.88}                & \textbf{7.84}                      \\ \hline
\bottomrule
\end{tabular}

\label{tab:fairfacecat}
\end{table*}
\subsection{UMAP Clustering Analysis and Influence of~$\alpha$} 
\begin{figure*}[htb]
\centering
\subfloat{\includegraphics[width=.31\textwidth]{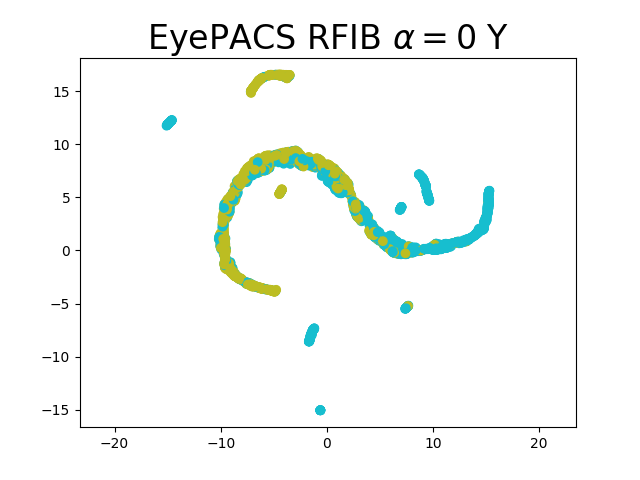}%
}
\subfloat{\includegraphics[width=.31\textwidth]{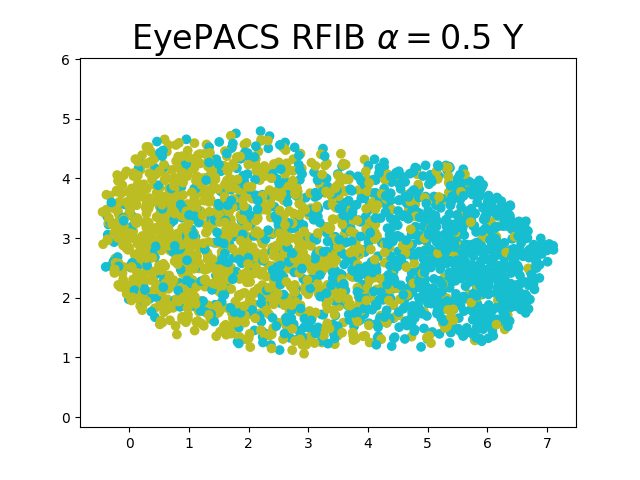}%
}
\subfloat{\includegraphics[width=.31\textwidth]{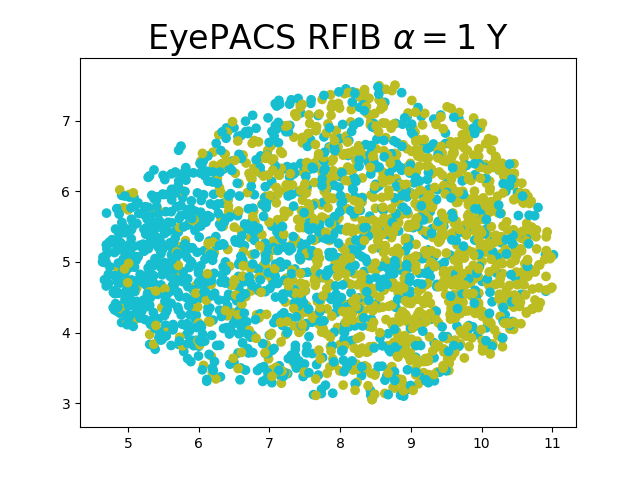}
}
\vspace{-0.1in}
\subfloat{\includegraphics[width=.31\textwidth]{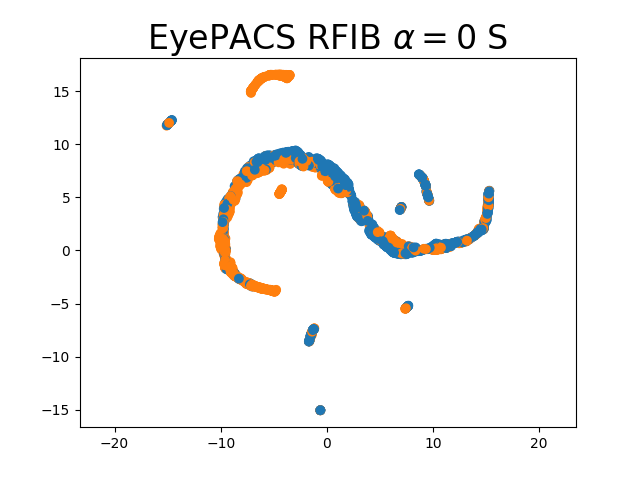}
}
\subfloat{\includegraphics[width=.31\textwidth]{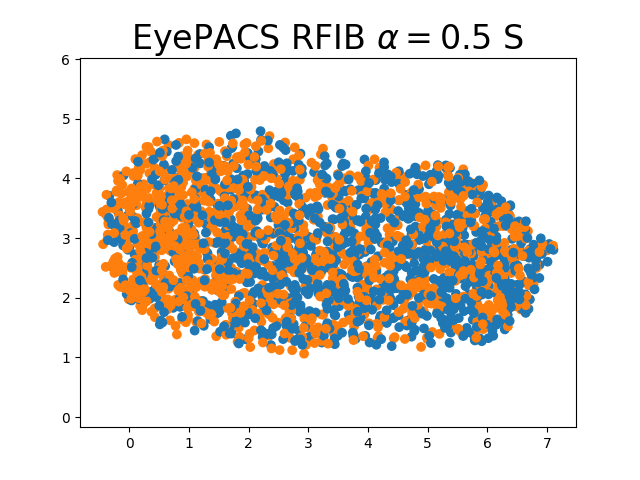}
}
\subfloat{\includegraphics[width=.31\textwidth]{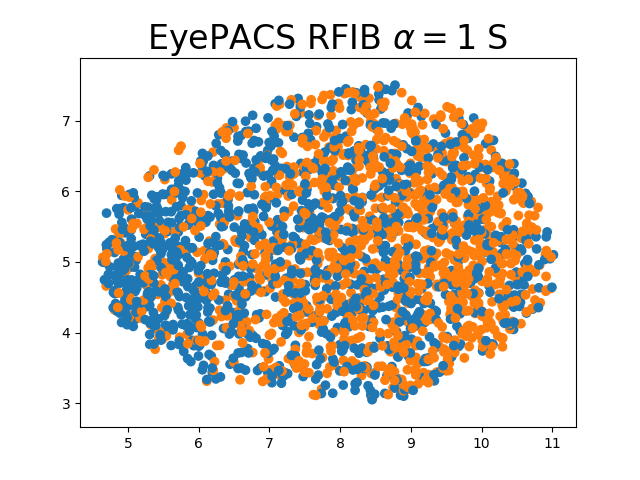}
}
\vspace{-0.1in}

\caption{UMAP dimensionality reduction of $Z$ for the EyePACS dataset using \Renyi divergence of order $\alpha$ in \eqref{final-loss}. The top row shows the two classes of $Y$ (light blue representing the positive examples and light green the negative) while the bottom row shows the two classes of $S$ (orange representing the positive examples and blue the negative). Going from left to right $\alpha$ is increased, resulting in more compression and causing a decrease in separation of the two classes. While separation occurs for both $Y$ and $S$, it occurs to a greater extent for $Y$ as desired.}
\label{fig:umap}
\end{figure*}

We use UMAP \cite{mcinnes2018umap} to provide a visual illustration of the effect of the \Renyi divergence on the RFIB system performance as it calibrates the compression term by varying its order $\alpha$. In Fig.~\ref{fig:umap}, we show 2-dimensional UMAP vectors of our representation $Z$, coloring the points both based on the label $Y$ and sensitive attribute $S$. Here experiments were done on the EyePACS dataset with $Y=$ DR and $S=$ ITA. The goal is for the representation to preserve information about $Y$, allowing the two classes of $Y$ to be easily separated, while removing the sensitive information $S$ and preventing its two classes from being distinguished.

A the value of $\alpha=0$ corresponds to no compression with $I(Z;X) = 0$, which preserves maximum accuracy but does not help fairness, with the classes of both $Y$ and $S$ being easily separated as shown in Fig.~\ref{fig:umap}. Increasing $\alpha$ gradually results in the different points getting more mixed together (recall that the \Renyi divergence is non-decreasing in its order $\alpha$). While this is seen for both $Y$ and $S$, the $I(Z;Y)$ and $I(Z;Y|S)$ terms help ensure that information about $Y$ is still preserved, allowing the classes of $Y$ to still separate fairly well even with higher compression, whereas the classes of $S$ get mixed together as desired.

We note that an intermediate value of $\alpha$ can potentially provide the best compromise between fairness and accuracy, as a more moderate amount of compression can be enough to sufficiently remove sensitive information and further compression might only harm accuracy. This is illustrated in Fig.~\ref{fig:umap} where a value of $\alpha = 0.5$ was sufficient to mix together the classes of $S$ while still preserving an obvious separation of the two classes of $Y$. Further increasing $\alpha$ worsened the separation of $Y$ more than it added additional benefit for $S$. This is also supported by our experimental results where best overall accuracy-fairness trade-offs were typically obtained for intermediate values of~$\alpha.$

\section{Conclusion}
We propose RFIB, a novel variational IB method based on \Renyi measures that offers trade-offs in utility, two fairness objectives, and compactness. Using \Renyi divergence and \Renyi cross-entropy instead of KL divergence gives a way to control the amount of compression with an additional hyper-parameter. Compared to prior work which incorporates a single definition of fairness, RFIB has the potential benefit of allowing ethicists and policy makers to specify softer and more balanced requirements for fairness that lie between multiple hard requirements, and our work opens the way to future studies expanding on this idea. Experimental results on three different image datasets and two tabular datasets showed RFIB provides benefits vis-a-vis
other methods of record including IB, CFB, and other techniques
performing augmentation or adversarial debiasing.

Possible directions of future research include using optimization methods other than the weighted sum method we used, such as the $\epsilon$-constraints method \cite{mavrotas}. They also include studying different techniques to minimize and maximize the mutual information terms, such as~\cite{wang2021}, the Mutual Information Neural Estimation algorithm \cite{mine} based on the Donsker–Varadhan representation of the Kullback-Leibler divergence~\cite{donsker1983asymptotic}, or the equivalent algorithm based on a variational representation of the \Renyi divergence~\cite{birrell2020}. In addition, we can further experiment with non-binary (categorical) targets, as outlined in Section \ref{sec:implementationDetails}, and study the performance of our method in cases where the sensitive attributes are not explicitly known. 

Finally, we note that our method is closely related to the problem of privacy where a privacy mechanism must be designed to release data that has high utility but does not reveal private information (e.g., see~\cite{sankar2013utility, asoodeh2016information, asoodeh2018estimation,rassouli2019optimal,issa2019operational,bloch2021overview, zamani2021data} and the references therein for various studies on information-theoretic privacy). While similar attributes can be used as sensitive or private variables, for fairness we are concerned with preventing the attribute from affecting classification accuracy, while for privacy we are concerned with preventing an attacker from accessing a user's private data. A very promising future extension of our work is to adapt our method to also remove private information in addition to sensitive information, achieving representations that are both private and fair. Similar to \cite{saeidian2021quantifying}, we can then quantify the amount of private information leakage and study theoretical guarantees on privacy for our method.

\balance
{\small
\bibliographystyle{IEEEtran}
\bibliography{egbib}
}

\end{document}